\documentclass{article}

\usepackage[preprint]{neurips_2025}

\usepackage[utf8]{inputenc} 
\usepackage[T1]{fontenc}    
\usepackage{hyperref}       
\usepackage{url}            
\usepackage{booktabs}       
\usepackage{amsfonts}       
\usepackage{amsmath}        
\usepackage{nicefrac}       
\usepackage{microtype}      
\usepackage{xcolor}         
\usepackage{graphicx}       
\usepackage{algorithm}
\usepackage{algpseudocode}
\usepackage{multirow}       
\usepackage{float}          

\title{Reasoning Topology Matters: Network-of-Thought for Complex Reasoning Tasks}

\author{%
  Fan Huang \\
  Indiana University Bloomington \\
  \texttt{huangfan@acm.org}
}

\begin{document}

\maketitle

\begin{abstract}
Existing prompting paradigms structure LLM reasoning in limited topologies: Chain-of-Thought (CoT) produces linear traces, while Tree-of-Thought (ToT) performs branching search. Yet complex reasoning often requires merging intermediate results, revisiting hypotheses, and integrating evidence from multiple sources. We propose Network-of-Thought (NoT), a framework that models reasoning as a directed graph with typed nodes and edges, guided by a heuristic-based controller policy. Across four benchmarks (GSM8K, Game of 24, HotpotQA, ProofWriter) and three models (GPT-4o-mini, Llama-3.3-70B-Instruct, Qwen2.5-72B-Instruct), we investigate when network topology outperforms chain or tree structures, whether LLM-generated heuristics can guide graph-based reasoning search, and the computation-accuracy tradeoff across topologies, evaluating each method on accuracy, topology simplicity, and token efficiency. Our results show that CoT remains effective for sequential tasks with GPT-4o-mini (89.5\% on GSM8K), while NoT surpasses ToT on multi-hop reasoning (91.0\% vs.\ 88.0\% on HotpotQA with LLM-as-Judge). With 72B open-source models, NoT achieves the highest accuracy on GSM8K (91.5\%), and Qwen2.5-72B achieves the best multi-hop QA result overall (91.7\% on HotpotQA). Self-generated controller heuristics outperform fixed and random strategies on logical reasoning, with uncertainty-only weighting achieving 57.0\% on ProofWriter. We also find that evaluation methodology significantly impacts method rankings: string-match underestimates all methods on open-ended QA, with the largest gap for NoT, a pattern consistent across all three models (14--18 percentage point gap on HotpotQA).
\end{abstract}

\vspace{-0.95em}

\begin{figure}[H]
\centering
\includegraphics[width=\textwidth]{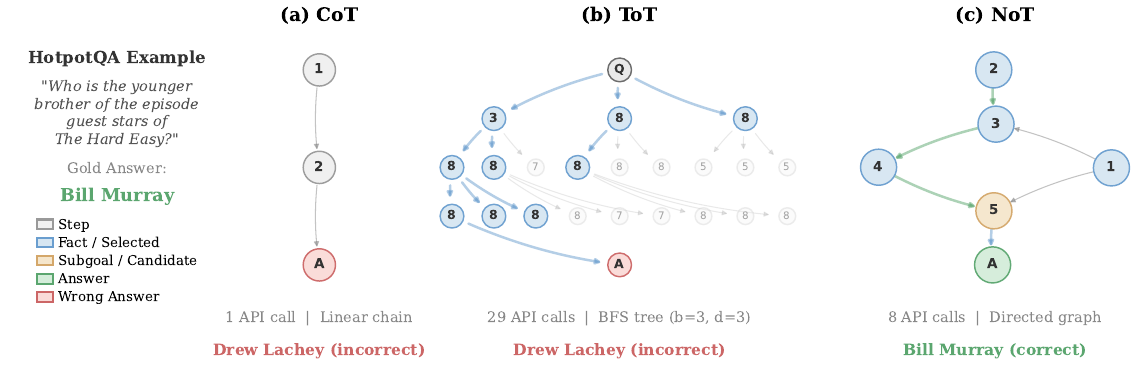}
\caption{Reasoning topologies on a HotpotQA~\citep{yang2018hotpotqa} multi-hop question. CoT and ToT both fixate on a single path (Nick Lachey $\to$ Drew Lachey), while NoT's graph structure pivots to the correct answer (Brian Doyle-Murray $\to$ Bill Murray) via multi-source dependency integration.}
\label{fig:highlight}
\end{figure}

\section{Introduction}
\label{sec:introduction}

Large language models have demonstrated strong reasoning capabilities through structured prompting methods~\citep{huang2023towards}. Chain-of-Thought (CoT)~\citep{wei2022chain} organizes reasoning as linear step-by-step traces, and subsequent work has extended this paradigm through least-to-most decomposition~\citep{zhou2023leasttomost}, complexity-based chain selection~\citep{fu2023complexity}, and self-consistency via majority voting~\citep{wang2023selfconsistency}. Tree-of-Thought (ToT)~\citep{yao2023tree} further extends reasoning to branching search over alternative paths. These methods impose specific \emph{reasoning topologies} on the reasoning process, yet the relationship between topology choice and task structure remains underexplored.

However, many complex decision tasks exhibit reasoning structures that are inherently \emph{network-like}. Multi-hop question answering requires integrating evidence from multiple independent sources into a single conclusion~\citep{yang2018hotpotqa}, and compositional reasoning exposes a systematic gap between single-hop and multi-hop accuracy that sequential decomposition alone does not close~\citep{press2023measuring,khot2023decomposed}. Logical reasoning involves hypothesis revision cycles where conclusions depend on multiple premises that may support or contradict each other~\citep{tafjord2021proofwriter}. Iterative refinement frameworks such as Self-Refine~\citep{madaan2023selfrefine} and Reflexion~\citep{shinn2023reflexion} demonstrate that cyclic feedback loops improve task performance, yet these operate at the output level rather than within a structured reasoning topology. These structural requirements cannot be naturally represented as a chain or tree.

We identify three fundamental limitations of existing reasoning topologies. First, the \emph{reasoning reuse problem}: when multiple reasoning paths converge on shared intermediate conclusions, trees must duplicate these nodes while graphs can merge them~\citep{besta2024graph}. Second, the \emph{reasoning loop problem}: hypothesis revision, iterative planning, and self-correction involve cyclic reasoning that chains and trees cannot express~\citep{madaan2023selfrefine,shinn2023reflexion}. Third, the \emph{multi-source dependency problem}: some decisions require integrating multiple simultaneous inputs into a single reasoning node, as demonstrated by multi-hop benchmarks that require evidence synthesis from disparate sources~\citep{yang2018hotpotqa,khot2023decomposed}.

We propose Network-of-Thought (NoT), which generalizes existing reasoning paradigms:
\[
\text{CoT} \subset \text{ToT} \subset \text{NoT} \qquad (\text{chain} \subset \text{tree} \subset \text{graph})
\]

The key contribution is not the graph structure itself, but the \emph{heuristic-guided controller policy} that operates on it. We introduce \emph{self-generated heuristics}, where the LLM itself proposes the controller weights that determine which reasoning node to expand next, replacing manual weight tuning with a single LLM call. Our experiments across four benchmarks reveal that while simple chain reasoning remains competitive for sequential tasks, network topology enables strong performance on multi-hop reasoning when paired with appropriate evaluation. We also find that the choice of evaluation methodology (string match vs. semantic judge) can shift relative method rankings by up to 18 percentage points, highlighting an important methodological consideration for reasoning research.

To systematically investigate these questions, we organize our study around three research questions: (RQ1)~When is network reasoning topology necessary over chain or tree structures? (RQ2)~Can self-generated heuristics improve network reasoning? (RQ3)~What is the computation-accuracy tradeoff across reasoning topologies?

Our contributions are:
\begin{enumerate}
    \item A reasoning topology taxonomy that formalizes the structural properties and tradeoffs of chain, tree, and network reasoning.
    \item The Network-of-Thought framework with a heuristic-guided controller policy for graph-structured reasoning with typed nodes and edges.
    \item Self-generated heuristics, where the LLM proposes the controller weights for node expansion scoring, replacing hand-tuned parameters with task-adaptive weight selection.
    \item Empirical analysis across four benchmarks and three model families (GPT-4o-mini, Llama-3.3-70B, Qwen2.5-72B) demonstrating when network topology helps, when it does not, and how evaluation methodology impacts the conclusions.
\end{enumerate}

We evaluate NoT along three dimensions: (1)~\textbf{accuracy} across diverse reasoning tasks,
(2)~\textbf{topology simplicity} measured by graph size and structure relative to tree-based search,
and (3)~\textbf{token efficiency} capturing the computation cost per correct answer. These three
metrics jointly assess whether graph-structured reasoning delivers meaningful gains over simpler
topologies at acceptable cost.

\begin{figure}[t]
\centering
\includegraphics[width=0.8\textwidth]{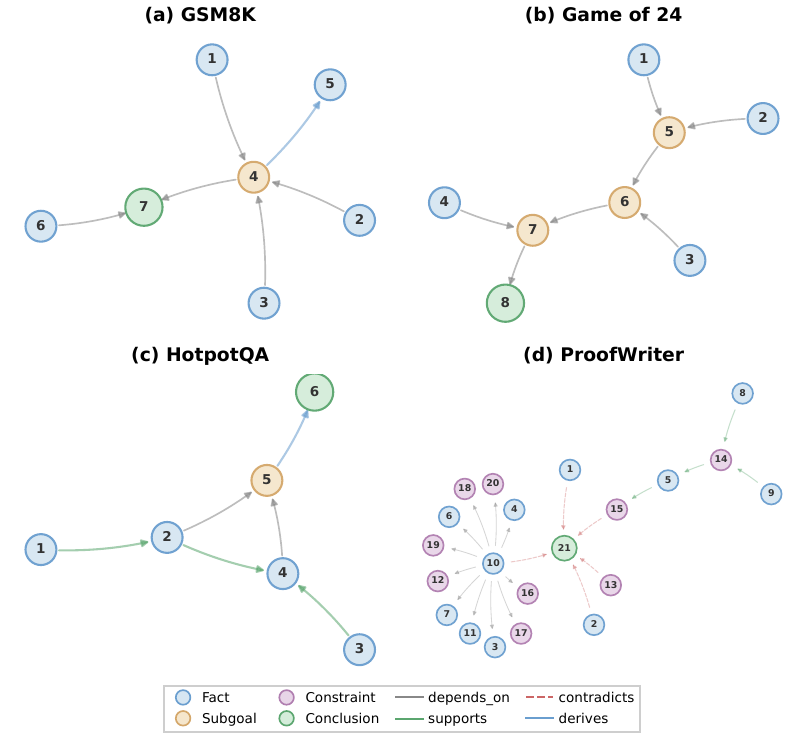}
\caption{Representative NoT reasoning graphs for each dataset. Node colors indicate type: blue (fact), orange (subgoal), pink (constraint), green (conclusion). Edge styles indicate relationship type. ProofWriter (d) produces the densest graph with 21 nodes reflecting its complex logical structure.}
\label{fig:not_graphs}
\end{figure}

\section{Related Work}
\label{sec:related_work}

\paragraph{Chain-of-Thought prompting.}
CoT~\citep{wei2022chain} decomposes reasoning into sequential intermediate steps, improving performance on arithmetic and commonsense tasks. Zero-shot CoT~\citep{kojima2022large} shows that the trigger phrase ``let's think step by step'' activates multi-step reasoning without in-context exemplars, while self-consistency~\citep{wang2023selfconsistency} improves reliability by sampling multiple independent chains and selecting the majority answer. Beyond the basic chain paradigm, least-to-most prompting~\citep{zhou2023leasttomost} decomposes complex problems into progressively simpler sub-questions solved sequentially, and complexity-based prompting~\citep{fu2023complexity} selects chains with more reasoning steps as exemplars, achieving stronger performance on multi-step tasks. Process-level verification~\citep{lightman2024lets} further improves chain reasoning by training reward models to evaluate each intermediate step rather than only the final answer. Despite these advances, all chain-based methods share a structural constraint: each reasoning trace forms a linear sequence, which precludes exploring alternative intermediate steps within a single trace or reusing shared sub-conclusions across traces.

\paragraph{Tree and graph reasoning topologies.}
ToT~\citep{yao2023tree} relaxes linearity by organizing reasoning into a branching search tree, where candidate continuations are proposed, evaluated, and pruned at each depth level via BFS or DFS. This enables deliberate exploration on planning tasks such as Game of 24~\citep{yao2023tree}. \citet{hao2023reasoning} adopt Monte Carlo Tree Search to guide tree-structured reasoning with a world model, and Algorithm of Thoughts~\citep{sel2024algorithm} explores in-context algorithmic search within a single LLM call. GoT~\citep{besta2024graph} further generalizes the topology to directed graphs, introducing aggregation, refinement, and generation operations over thought units, while Everything of Thoughts~\citep{ding2024everything} combines MCTS with graph-based thought representations. However, these approaches rely on hand-designed transformation rules and evaluate each topology in isolation without systematic comparison across problem types or analysis of when graph structure is necessary versus when simpler topologies suffice.

\paragraph{Iterative refinement and cyclic reasoning.}
Self-Refine~\citep{madaan2023selfrefine} demonstrates that LLMs can iteratively critique and revise their own outputs through feedback loops, improving performance across code generation, math reasoning, and sentiment reversal tasks. Reflexion~\citep{shinn2023reflexion} extends this idea to language agents that store verbal reflections on past failures and use them to improve subsequent attempts. CRITIC~\citep{gou2024critic} introduces tool-interactive critiquing, where LLMs verify and correct their outputs by interacting with external tools such as search engines and code interpreters in a generate-critique-revise loop. \citet{xie2023selfeval} propose self-evaluation guided beam search, in which the LLM scores its own intermediate reasoning steps to prune unpromising paths during multi-step inference. Language Agent Tree Search~\citep{zhou2024language} unifies reasoning, acting, and planning by integrating Monte Carlo Tree Search with LLM-powered self-reflections, enabling iterative search-and-reflect cycles. These methods implement various forms of cyclic reasoning through repeated self-evaluation, but operate at the output level or within fixed search procedures rather than within a structured reasoning graph. Our framework integrates such revision cycles directly into the graph topology through typed edges and controller-guided re-expansion.

\paragraph{LLM-based evaluation.}
\citet{zheng2023judging} show that LLM-based evaluation provides more reliable assessment than string matching when reference answers admit multiple valid surface forms. \citet{liu2023geval} demonstrate that GPT-4-based evaluation achieves higher correlation with human judgments than traditional automatic metrics for text generation, and \citet{chiang2023can} confirm that LLM evaluators can serve as a cost-effective proxy for human evaluation across multiple NLP tasks. Prometheus~\citep{kim2024prometheus} induces fine-grained evaluation capability in open-source language models by training on score rubrics, enabling reproducible evaluation without proprietary API dependence. AlpacaFarm~\citep{dubois2023alpacafarm} provides a simulation framework that validates LLM-based automatic evaluators against human annotations for instruction-following tasks. This methodological concern is directly relevant to topology comparison: an overly strict metric may penalize methods that produce semantically correct but lexically divergent answers, distorting the relative ranking of reasoning topologies. We adopt both string match and LLM-as-Judge to quantify how evaluation methodology affects topology comparisons.

\paragraph{Tool-augmented reasoning.}
ReAct~\citep{yao2022react} interleaves reasoning traces with external tool-use actions, grounding language model reasoning in retrieved information. Toolformer~\citep{schick2023toolformer} trains language models to autonomously decide when and how to invoke external tools, while PAL~\citep{gao2023pal} delegates computation to a program interpreter, separating reasoning from execution. Decomposed prompting~\citep{khot2023decomposed} routes sub-problems to specialized modules, combining tool use with task decomposition. Tool augmentation is orthogonal to reasoning topology, as action nodes can be incorporated within any graph structure. We focus on the topology dimension rather than external tool integration.

\section{Reasoning Topology Taxonomy}
\label{sec:taxonomy}

We propose a taxonomy of reasoning topologies based on their structural properties.

\begin{table}[t]
\centering
\caption{Reasoning topology taxonomy. Each topology offers a different tradeoff between computational cost and expressiveness.}
\label{tab:taxonomy}
\begin{tabular}{@{}llll@{}}
\toprule
\textbf{Topology} & \textbf{Structure} & \textbf{Reasoning property} & \textbf{Compute cost} \\
\midrule
Chain (CoT) & Linear & Sequential reasoning & $O(d)$ \\
Tree (ToT) & Branching & Search reasoning & $O(b^d)$ \\
Network (NoT) & Directed graph & Dependency reasoning & $O(|V| + |E|)$ \\
\bottomrule
\end{tabular}
\end{table}

Each topology has a computation-expressiveness tradeoff (Table~\ref{tab:taxonomy}): chains are cheap but limited, trees are expensive but enable search, and networks offer high expressiveness at moderate cost but with higher maintenance overhead.

\subsection{Limitations of chain and tree topologies}

We formalize three limitations of existing reasoning topologies that motivate the need for network structures.

\paragraph{Reasoning reuse problem.}
Given reasoning paths $A \rightarrow B \rightarrow D$ and $A \rightarrow C \rightarrow D$, tree structures must duplicate node $D$. Graph structures share it, reducing redundancy. Our experiments show that NoT reasoning graphs exhibit reuse rates of 0.08 to 0.28 across datasets (Table~\ref{tab:rq3}), confirming that shared intermediate conclusions arise naturally.

\paragraph{Reasoning loop problem.}
Iterative reasoning processes (hypothesis $\rightarrow$ test $\rightarrow$ revise $\rightarrow$ hypothesis) form cycles. Chains and trees are acyclic by definition and cannot represent such patterns. While our current experiments show zero loop frequency (the LLM does not spontaneously generate cyclic structures), this limitation motivates future work on loop-encouraging prompting strategies.

\paragraph{Multi-source dependency problem.}
When a reasoning conclusion depends on multiple independent inputs simultaneously, chain topology is inadequate and tree topology is unnatural. Graph topology handles multi-input nodes directly. Our HotpotQA results support this: NoT builds graphs with average dependency complexity of 0.86 edges per node, reflecting the multi-source nature of multi-hop question answering.

\subsection{Reasoning graph representation}

A Network-of-Thought reasoning process is represented as a directed graph $G = (V, E)$, where each node $v \in V$ represents a typed reasoning unit and each edge $(u, v) \in E$ represents a typed semantic relationship between reasoning units.

\paragraph{Node types.} Each node has a type drawn from the set \{\texttt{fact}, \texttt{subgoal}, \texttt{constraint}, \texttt{conclusion}\}, along with a confidence score $c_v \in [0, 1]$, a status (unresolved, resolved, or contradicted), and natural language content. The type taxonomy is specified per task in the initialization prompt, though the LLM is not hard-constrained to these types. Across all experiments, 99.8\% of generated nodes use the prompted types, with occasional LLM drift producing types such as \texttt{calculation} (1.4\%) or \texttt{hypothesis} (0.4\%).

\paragraph{Edge types.} Each edge has a type from \{\texttt{depends\_on}, \texttt{supports}, \texttt{derives}, \texttt{contradicts}\} and a weight $w_e \in [0, 1]$. Edge types are similarly specified per task, with 99.8\% adherence across all experiments. The \texttt{contradicts} type is specific to logical reasoning tasks (ProofWriter), while the other three types appear across all task categories.

Unlike trees, this representation naturally supports shared nodes (reasoning reuse), multi-input nodes (dependency reasoning), and typed relationships that distinguish supporting from contradicting evidence.

\subsection{Models}

Our primary experiments use GPT-4o-mini~\citep{openai2024gpt4o} with temperature 0.1 and seed 42 for reproducibility. To test cross-model generalization, we replicate the full experimental pipeline with two open-source 72B models: Llama-3.3-70B-Instruct~\citep{grattafiori2024llama} and Qwen2.5-72B-Instruct~\citep{yang2024qwen25}. We compare three reasoning topologies: Chain-of-Thought (CoT), which uses a single LLM call with ``think step by step'' prompting (1 API call per instance); Tree of Thoughts (ToT), which performs BFS with branching factor $b=3$ and maximum depth $d=3$, executing propose--evaluate--select phases at each depth level (29 API calls per instance); and Network of Thoughts (NoT), which performs graph-based reasoning with controller-guided expansion (3--22 API calls per instance, averaging 12, depending on convergence speed).

\subsection{Evaluation metrics}
\label{sec:eval_metrics}

To compare reasoning topologies systematically, we evaluate along three core dimensions:

\paragraph{Topology simplicity.} We measure the structural complexity of each reasoning topology by graph size (number of nodes $|V|$ and edges $|E|$) and dependency complexity (edges per node). Simpler topologies are preferable when they achieve comparable accuracy, as they are easier to interpret and less prone to error propagation. A linear chain has minimal complexity ($|V| = d$, $|E| = d{-}1$), while a BFS tree grows exponentially ($O(b^d)$ nodes). Network topologies fall between these extremes, producing compact graphs that capture multi-source dependencies without redundant branching.

\paragraph{Reasoning accuracy.} We measure task-level correctness using both string-match and semantic (LLM-as-Judge) evaluation. String-match provides reproducible scoring but underestimates methods that produce semantically correct but lexically different answers. LLM-as-Judge evaluation better captures reasoning quality, particularly for open-ended tasks where multiple valid phrasings exist.

\paragraph{Token efficiency.} We quantify the computation cost of each topology as total tokens consumed per correct answer. This metric captures the practical cost of reasoning: a method that achieves high accuracy but requires disproportionate computation may be impractical. Token efficiency jointly reflects prompt overhead, the number of LLM calls, and the success rate.

These three metrics provide a unified framework for assessing whether more expressive topologies deliver meaningful gains over simpler alternatives at acceptable cost. Beyond these dimensions, reasoning graphs open avenues for deeper structural analysis. Future work may leverage network-theoretic tools such as node centrality, community detection, and cluster analysis to reveal which reasoning substructures drive correct conclusions and how information flows through complex reasoning processes.

In practice, we operationalize accuracy with two complementary scoring methods. \textbf{String match (SM)} applies task-specific exact match after normalization: numeric extraction for GSM8K, expression evaluation for Game of 24, normalized string match for HotpotQA, and label match for ProofWriter. \textbf{LLM-as-Judge (Judge)} uses GPT-4o-mini to evaluate whether the prediction and gold answer are semantically equivalent; items passing string match are auto-judged correct and empty predictions are auto-judged incorrect, with results cached for reproducibility. This evaluation is applied to all methods on all datasets for fair comparison. We report CoT and ToT with string match, and NoT with both string match and LLM-as-Judge, reflecting the observation that string match disproportionately penalizes NoT's more verbose extraction while the LLM-as-Judge provides a fairer assessment of semantic correctness.

\section{Network-of-Thought Framework}
\label{sec:framework}

\subsection{Framework pipeline}

\begin{figure*}[!htbp]
\centering
\includegraphics[width=0.95\textwidth]{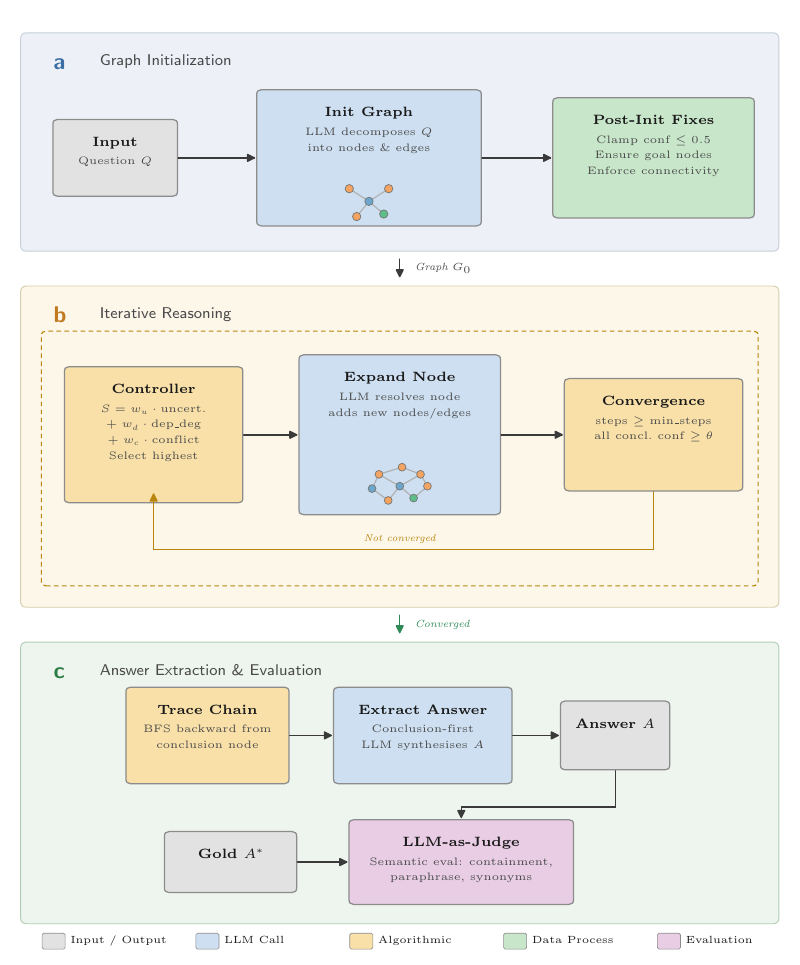}
\caption{NoT framework pipeline. (a)~\textbf{Graph Initialization}: the input problem is converted into a typed reasoning graph by an LLM call, then post-processed to clamp confidence, ensure goal nodes, and enforce connectivity. (b)~\textbf{Iterative Reasoning}: the controller scores unresolved nodes using configurable weights $w_u$, $w_d$, $w_c$ (Section~\ref{sec:controller}), the highest-priority node is expanded via an LLM call, and convergence is checked (all conclusion confidences $\geq \theta$); if not converged, control loops back to the controller. (c)~\textbf{Answer Extraction \& Evaluation}: a backward BFS traces the reasoning chain, a conclusion-first prompt extracts the final answer, and an LLM-as-Judge evaluates semantic equivalence against the gold answer.}
\label{fig:pipeline}
\end{figure*}

The NoT framework operates in three stages (Figure~\ref{fig:pipeline}a--c), comprising six phases. In Phase~1 (panel~a), a single LLM call analyzes the input problem and produces an initial reasoning graph with typed nodes and edges in JSON format, using task-specific initialization prompts that specify the expected node and edge types for each domain (full prompts in Appendix~\ref{app:prompts}). The resulting graph then undergoes post-initialization fixes: confidence scores are clamped to $\leq 0.5$ because the LLM tends to assign near-perfect confidence at initialization, causing immediate convergence before any iterative refinement occurs; goal nodes are ensured to exist because certain task domains (e.g., ProofWriter) frequently produce graphs without conclusion nodes, leading to convergence loops that run to the maximum step limit; and connectivity is enforced by adding bridge edges between disconnected components, since isolated subgraphs prevent reasoning from propagating across the full problem structure. In Phase~2 (panel~b), the controller computes a priority score for each unresolved node (Section~\ref{sec:controller}) and selects the highest-scoring node for expansion, combining uncertainty, dependency degree, and conflict signals with configurable weights. Phase~3 expands the selected node via a single LLM call that provides updated confidence, new nodes, and new edges, with the expansion prompt including the full graph context and all resolved node values. Phase~4 checks whether all conclusion nodes have reached confidence $\geq \theta$ (with $\theta = 0.8$) and at least 3 expansion steps have been completed; if not converged, control returns to Phase~2 for another iteration, and the loop also terminates if a maximum of 20 expansion steps is reached or no confidence change exceeds 0.05 in the last two steps (stagnation detection). After convergence (panel~c), Phase~5 performs a backward BFS from conclusion nodes to trace the reasoning chain through subgoals back to supporting facts, producing an ordered reasoning path. Finally, Phase~6 extracts the answer using a conclusion-first strategy: the extraction prompt anchors on the best conclusion node's value, guiding the model to produce a final answer grounded in the traced reasoning chain. The predicted answer is then evaluated by an LLM-as-Judge that checks semantic equivalence against predefined criteria (containment, paraphrase, and synonym matching).

\subsection{Dataset selection rationale}
\label{sec:datasets}

\begin{table}[t]
\centering
\caption{Datasets and expected topology-task matching.}
\label{tab:datasets}
\small
\begin{tabular}{@{}llrl@{}}
\toprule
\textbf{Dataset} & \textbf{Task type} & \textbf{$N$} & \textbf{Expected best topology} \\
\midrule
GSM8K~\citep{cobbe2021gsm8k} & Math reasoning & 200 & Chain (sequential arithmetic) \\
Game of 24 & Search reasoning & 100 & Tree (combinatorial search) \\
HotpotQA~\citep{yang2018hotpotqa} & Multi-hop QA & 300 & Network (multi-source) \\
ProofWriter~\citep{tafjord2021proofwriter} & Logical reasoning & 300 & Network (hypothesis revision) \\
\bottomrule
\end{tabular}
\end{table}

Our evaluation requires datasets that collectively span the three reasoning topologies in our taxonomy (Section~\ref{sec:taxonomy}), so that we can test whether topology-task matching holds in practice. We select four benchmarks (Table~\ref{tab:datasets}), each targeting a different structural requirement: GSM8K~\citep{cobbe2021gsm8k} (200 instances) serves as our chain-topology control, where sequential arithmetic forms a linear dependency chain and any benefit from graph structure would indicate unnecessary complexity; Game of 24 (100 instances, following the protocol of \citet{yao2023tree}) targets tree-topology reasoning through combinatorial search over operator and operand pairings; HotpotQA~\citep{yang2018hotpotqa} (300 instances, bridge type) targets network-topology reasoning through multi-source dependency, requiring synthesis of evidence from two independent documents into a single conclusion; and ProofWriter~\citep{tafjord2021proofwriter} (300 instances, open-world, depth-5) targets network-topology reasoning through hypothesis revision, where logical deduction with negation demands tracking supporting and contradicting evidence simultaneously via typed edges.

Sample sizes balance statistical power against the computational cost of multi-call methods: NoT requires 5--20 LLM calls per instance, making large-scale evaluation expensive. We allocate more instances to datasets with higher answer variability: HotpotQA (300) for open-ended QA with greater format variance, ProofWriter (300) for adequate coverage of its three-class output space (True, False, Unknown), and GSM8K (200) for low-variance numeric answers. The total of 900 instances yields approximately 27,000 LLM calls across the three-method comparison, providing sufficient resolution to detect accuracy differences of 5 percentage points or greater. Together, these four datasets provide a controlled comparison: two tasks where simpler topologies should suffice (GSM8K, Game of 24) and two where graph structure should provide genuine advantages (HotpotQA, ProofWriter), allowing us to test not only whether NoT works but also when it is and is not beneficial relative to simpler alternatives.

\subsection{Controller policy and self-generated heuristics}
\label{sec:controller}

The controller policy selects which node to expand next from the set of unresolved nodes:
\begin{equation}
\pi(v \mid G) = \arg\max_{v \in V_{\text{unresolved}}} \; \text{score}(v)
\end{equation}

where the scoring function combines three heuristic signals:
\begin{equation}
\text{score}(v) = w_u \cdot \text{uncertainty}(v) + w_d \cdot \text{dep\_degree}(v) + w_c \cdot \text{conflict}(v)
\label{eq:score}
\end{equation}

The uncertainty component is $1 - c_v$, where $c_v$ is the node's confidence. The dependency degree counts the fraction of unresolved successor nodes that would be unblocked by resolving $v$. The conflict component is 1 if any edge of type \texttt{contradicts} touches $v$, and 0 otherwise.

The weights $w_u$, $w_d$, $w_c$ shown in the controller box of Figure~\ref{fig:pipeline} are configurable; we set the defaults to $w_u = 0.4$, $w_d = 0.4$, $w_c = 0.2$. This allocation draws on two established principles. First, uncertainty sampling is a well-studied acquisition function in active learning~\citep{settles2012active, lewis1994sequential}, where selecting the least-certain instance for labeling consistently outperforms random selection; we adapt this by prioritizing the least-confident reasoning node for expansion. Second, dependency-driven scheduling follows the critical-path heuristic from planning and search~\citep{pearl1984heuristics}, where resolving bottleneck nodes that unblock the most downstream work maximizes progress per step. We assign equal weight to these two complementary signals ($w_u = w_d = 0.4$) because uncertainty identifies \emph{what} needs refinement while dependency degree identifies \emph{what} would be most impactful to resolve. The conflict component receives lower weight ($w_c = 0.2$) because contradictions are sparse; they arise primarily in logical reasoning tasks (ProofWriter) and are absent in arithmetic and QA domains, so it serves as a tie-breaking signal rather than a primary driver.

Rather than requiring these weights to be hand-tuned per task, we also explore a self-generated heuristic variant: we prompt the LLM to propose the weights $(w_u, w_d, w_c)$ by describing the graph structure (node types with confidence scores, edge types with weights) and asking it to output values for the three scoring components. The LLM is called \emph{once} before all instances in a given experiment; the resulting weights are then used as a fixed scoring function throughout the run. In our experiments, the LLM proposed weights $(0.5, 0.3, 0.2)$, placing higher emphasis on uncertainty than the hand-designed default. This lightweight mechanism replaces manual weight tuning with a single LLM call, enabling task-adaptive controller configuration without per-domain engineering.

\section{Results}
\label{sec:results}

\subsection{RQ1: When is network reasoning topology necessary?}

\begin{table}[t]
\centering
\caption{Main accuracy results (\%) across four benchmarks. CoT and ToT are evaluated by string match (SM); NoT is evaluated by both string match and LLM-as-Judge. Bold indicates best per dataset. Token costs are reported in Table~\ref{tab:rq3}.}
\label{tab:main_results}
\small
\begin{tabular}{l cc cc}
\toprule
\textbf{Dataset} & \textbf{CoT (SM)} & \textbf{ToT (SM)} & \textbf{NoT (SM)} & \textbf{NoT (Judge)} \\
\midrule
GSM8K & \textbf{89.5} & 69.5 & 82.0 & 85.0 \\
Game of 24 & 58.0 & 11.0 & 75.0 & \textbf{86.0} \\
HotpotQA & 86.3 & 72.3 & 73.3 & \textbf{91.0} \\
ProofWriter & 43.3 & \textbf{51.7} & 49.0 & 50.3 \\
\bottomrule
\end{tabular}
\end{table}

\begin{figure}[t]
\centering
\includegraphics[width=\textwidth]{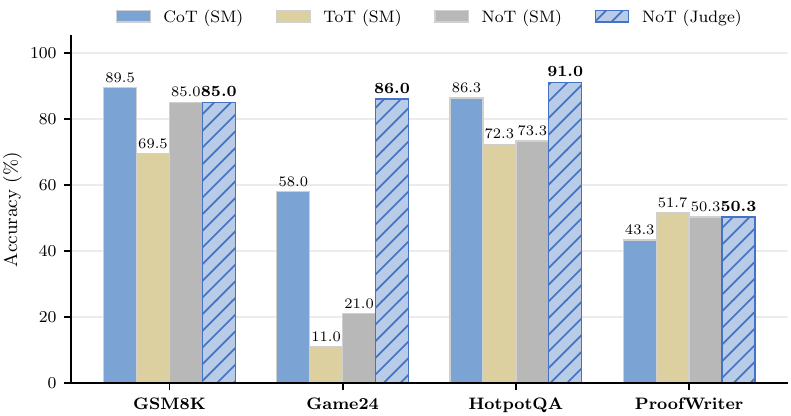}
\caption{Main accuracy comparison across four benchmarks. CoT and ToT evaluated by string match; NoT shown with both string match (gray) and LLM-as-Judge (hatched). The hatched bar represents the primary NoT evaluation.}
\label{fig:main_accuracy}
\end{figure}

Table~\ref{tab:main_results} presents the main accuracy results. Figure~\ref{fig:main_accuracy} provides a visual comparison.

\paragraph{GSM8K: Chain topology suffices.} CoT achieves the highest accuracy at 89.5\%, confirming that sequential math reasoning does not benefit from more complex topologies. NoT reaches 85.0\% (Judge), a competitive result but 4.5 percentage points below CoT, while using 15.8 times more tokens (7,748 vs. 490). The network overhead is not justified for this task type.

\paragraph{Game of 24: Evaluation reveals ToT weakness.} ToT achieves only 11.0\% accuracy under honest evaluation (after removing the \texttt{eval("24")==24} loophole where bare ``24'' predictions trivially pass). CoT achieves 58.0\% with genuine arithmetic expressions. NoT reaches 86.0\% under Judge evaluation, substantially outperforming both baselines on this task.

\paragraph{HotpotQA: NoT competitive on multi-hop reasoning.} NoT achieves 91.0\% under Judge evaluation, surpassing ToT (72.3\% SM) and approaching CoT (86.3\% SM). When all methods are evaluated with LLM-as-Judge, NoT (91.0\%) surpasses ToT (88.0\%) by 3.0 percentage points but trails CoT (95.7\%) by 4.7 percentage points. The string-match metric underestimates NoT by 17.7 percentage points (73.3\% SM vs. 91.0\% Judge), the largest gap among all methods.

\paragraph{ProofWriter: ToT edges out NoT.} ToT achieves the highest accuracy at 51.7\% (SM), followed by NoT at 50.3\% (Judge) and CoT at 43.3\%. The margin between ToT and NoT is narrow (1.4 percentage points). ProofWriter's logical reasoning with negation under open-world assumption remains challenging for all methods.

\paragraph{Summary.} The hypothesis that network topology is necessary for multi-hop and logical reasoning is partially supported. CoT remains the strongest method for sequential tasks. NoT shows its best relative performance on HotpotQA, where multi-source dependency reasoning benefits from graph structure. However, the gap to CoT suggests that the current NoT framework does not yet fully exploit the potential of network-structured reasoning.

\subsection{RQ2: Can self-generated heuristics improve network reasoning?}

\begin{table}[t]
\centering
\caption{Controller strategy comparison and weight ablation results on the two network-favoring datasets (Table~\ref{tab:datasets}): HotpotQA, where NoT shows its largest accuracy gain over baselines, and ProofWriter, where NoT shows its smallest gain, together spanning the full range of NoT effectiveness. SM = string match, Judge = LLM-as-Judge. Bold indicates best per section.}
\label{tab:rq2}
\small
\begin{tabular}{ll cc c}
\toprule
\textbf{Dataset} & \textbf{Controller} & \textbf{SM (\%)} & \textbf{Judge (\%)} & \textbf{Avg Tokens} \\
\midrule
HotpotQA & Random & 73.3 & \textbf{88.7} & 12,215 \\
 & Fixed & 71.7 & 88.0 & 14,971 \\
 & Self-Gen. & 71.7 & 88.0 & 15,068 \\
\addlinespace
ProofWriter & Random & 49.7 & 49.7 & 30,906 \\
 & Fixed & 51.3 & 51.3 & 28,411 \\
 & Self-Gen. & \textbf{54.0} & \textbf{54.0} & 26,536 \\
\midrule
\multicolumn{5}{l}{\textit{Weight ablation (ProofWriter, fixed controller)}} \\
\midrule
 & $w_u$ only & \textbf{57.0} & \textbf{57.0} & 29,560 \\
 & $w_d$ only & 51.0 & 51.0 & 35,127 \\
 & $w_c$ only & 49.0 & 49.0 & 29,549 \\
 & $w_u + w_d$ & 52.3 & 52.3 & 29,686 \\
 & Full ($w_u{+}w_d{+}w_c$) & 49.7 & 49.7 & 28,310 \\
\bottomrule
\end{tabular}
\end{table}

\begin{figure}[t]
\centering
\includegraphics[width=\textwidth]{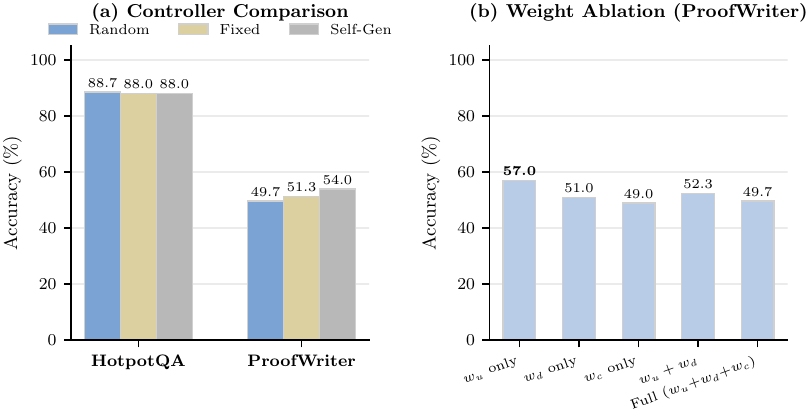}
\caption{Controller strategy comparison (left) and weight ablation on ProofWriter (right). Self-generated heuristics (54.0\%) outperform both the hand-designed fixed controller (51.3\%) and random baseline (49.7\%) on ProofWriter. Uncertainty-only weighting ($w_u$) achieves the best ablation result at 57.0\%.}
\label{fig:controller}
\end{figure}

Table~\ref{tab:rq2} presents the controller comparison and weight ablation results. Figure~\ref{fig:controller} provides a visual comparison.

\paragraph{Controller comparison.} On ProofWriter, self-generated heuristics achieve the highest accuracy (54.0\%), outperforming the hand-designed fixed controller (51.3\%) by 2.7 percentage points and the random baseline (49.7\%) by 4.3 percentage points. Notably, the only difference between the self-generated and fixed controllers is the weight values: the LLM proposed $(0.5, 0.3, 0.2)$ versus the hand-tuned default $(0.4, 0.4, 0.2)$, shifting emphasis toward uncertainty. This single-call weight generation, requiring no domain-specific examples or iterative tuning, produces a measurably better controller for logical reasoning. The self-generated controller also uses fewer tokens on ProofWriter (26,536 vs. 30,906 for random), indicating more efficient reasoning paths. On HotpotQA, controller differences are minimal under Judge evaluation (88.0\%--88.7\%), suggesting that for multi-hop QA the graph structure itself matters more than node selection order.

\paragraph{Weight ablation.} On ProofWriter, uncertainty-only weighting ($w_u = 1.0$, others zero) achieves the highest accuracy at 57.0\%, outperforming the full three-component scoring (49.7\%) by 7.3 percentage points. This finding indicates that for logical reasoning, prioritizing the most uncertain nodes is more effective than balancing multiple heuristic signals. The dependency-only component ($w_d$) is second best at 51.0\%, while the conflict component ($w_c$) alone achieves only 49.0\%. The full combination underperforms the individual components, suggesting potential interference between heuristic signals.

\subsection{RQ3: What is the computation-accuracy tradeoff?}

\begin{table}[t]
\centering
\caption{Computation cost and graph topology across methods and datasets. CoT and ToT evaluated by string match (SM); NoT evaluated by LLM-as-Judge (J). Tokens/Correct = average tokens divided by accuracy fraction. Nodes and Edges are averages per instance. Density $= |E|/(|V|(|V|{-}1))$ is the directed graph density. For topology comparison, we operationalize CoT nodes as implicit reasoning steps counted from numbered items in the output trace (edges $=$ nodes $-$ 1 for the linear chain; see Appendix~\ref{app:cot_steps}); ToT is a fixed BFS tree ($b{=}3$, $d{=}3$, 21 nodes, 20 edges); NoT nodes and edges are extracted from the generated reasoning graph.}
\label{tab:rq3}
\small
\begin{tabular}{ll rrr rr r}
\toprule
\textbf{Dataset} & \textbf{Method} & \textbf{Acc (\%)} & \textbf{Avg Tok} & \textbf{Tok/Corr} & \textbf{Avg Nodes} & \textbf{Avg Edges} & \textbf{Avg Density} \\
\midrule
GSM8K & CoT (SM) & 89.5 & 490 & 548 & 7.3 & 6.3 & 0.137 \\
 & ToT (SM) & 69.5 & 6,347 & 9,132 & 21 & 20 & 0.048 \\
 & NoT (J) & 85.0 & 7,748 & 9,116 & 7.2 & 8.8 & 0.196 \\
\addlinespace
Game of 24 & CoT (SM) & 58.0 & 1,161 & 2,002 & 4.1 & 3.1 & 0.244 \\
 & ToT (SM) & 11.0 & 6,352 & 57,745 & 21 & 20 & 0.048 \\
 & NoT (J) & 86.0 & 16,004 & 18,609 & 9.2 & 10.4 & 0.139 \\
\addlinespace
HotpotQA & CoT (SM) & 86.3 & 1,537 & 1,781 & 3.7 & 2.7 & 0.270 \\
 & ToT (SM) & 72.3 & 44,298 & 61,270 & 21 & 20 & 0.048 \\
 & NoT (J) & 91.0 & 15,254 & 16,763 & 4.9 & 4.2 & 0.220 \\
\addlinespace
ProofWriter & CoT (SM) & 43.3 & 562 & 1,297 & 5.1 & 4.1 & 0.196 \\
 & ToT (SM) & 51.7 & 9,043 & 17,491 & 21 & 20 & 0.048 \\
 & NoT (J) & 50.3 & 27,096 & 53,833 & 19.5 & 19.0 & 0.052 \\
\bottomrule
\end{tabular}
\end{table}

\begin{figure}[t]
\centering
\includegraphics[width=\textwidth]{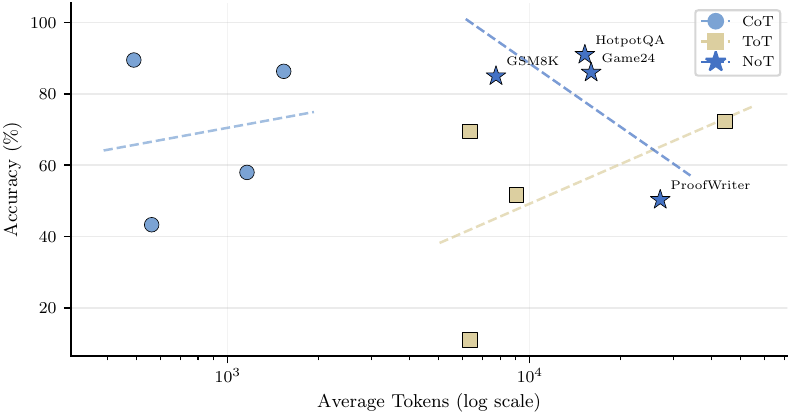}
\caption{Accuracy vs. computation (average tokens) across methods and datasets. Each point represents a method-dataset combination. CoT consistently occupies the low-token, high-accuracy region. NoT uses moderate tokens with competitive accuracy on multi-hop tasks.}
\label{fig:computation}
\end{figure}

Table~\ref{tab:rq3} presents the computation cost and graph topology metrics. Figure~\ref{fig:computation} shows the accuracy-computation relationship.

\paragraph{Token efficiency.} CoT is the most token-efficient method across all datasets, achieving the lowest tokens-per-correct-answer ratio (547 on GSM8K, 1,781 on HotpotQA, 1,298 on ProofWriter). This is expected given that CoT uses a single API call per instance. NoT uses moderate tokens (7,748 to 27,096 average) and achieves substantially better token efficiency than ToT on HotpotQA (16,763 vs.\ 61,270 tokens per correct answer) and Game of 24 (18,609 vs.\ 57,745). ToT is the most token-expensive method, particularly on HotpotQA where its BFS exploration results in 44,298 average tokens.

\paragraph{Topology simplicity.} The three methods span a wide range of structural complexity. CoT produces linear chains of 3.7--7.3 implicit reasoning steps depending on task complexity, with the simplest topology overall. ToT generates a fixed BFS tree with 21 nodes and 20 edges regardless of task, yielding a uniformly low graph density of 0.05. NoT produces adaptive graphs whose size and density vary by task: compact but relatively dense graphs on HotpotQA (4.9 nodes, density 0.22) that capture multi-source dependencies without redundant branching, moderate graphs on GSM8K (7.2 nodes, density 0.20) and Game of 24 (9.2 nodes, density 0.14), and large but sparse graphs on ProofWriter (19.5 nodes, density 0.05) whose tree-like structure resembles ToT despite being generated adaptively. Notably, NoT's HotpotQA graphs use fewer than a quarter of ToT's nodes while achieving higher accuracy and greater connectivity.

\paragraph{Computation-accuracy tradeoff.} The relationship between computation and accuracy is method-dependent. For CoT, increased token usage (from different task complexities) correlates positively with accuracy. For NoT, the relationship is more nuanced: HotpotQA achieves the best NoT accuracy (91.0\%) at moderate token cost (15,254), while ProofWriter uses the most tokens (27,096) with lower accuracy (50.3\%). This suggests that graph-structured reasoning is most efficient when the task naturally requires multi-source integration, but becomes less efficient when the logical complexity exceeds the model's graph-building capabilities.

\subsection{Cross-model generalization}
\label{sec:cross_model}

To test whether our findings generalize beyond GPT-4o-mini, we replicate the full experimental pipeline with Llama-3.3-70B-Instruct and Qwen2.5-72B-Instruct. The open-source models use the same prompts, controller settings, and evaluation procedures, with identical dataset sizes across all three models.

\begin{table}[t]
\centering
\caption{Cross-model accuracy comparison (\%). CoT and ToT evaluated by string match (SM); NoT evaluated by both SM and LLM-as-Judge (J). Bold indicates best per dataset-model combination.}
\label{tab:cross_model}
\small
\begin{tabular}{@{}ll cccc@{}}
\toprule
\textbf{Dataset} & \textbf{Model} & \textbf{CoT (SM)} & \textbf{ToT (SM)} & \textbf{NoT (SM)} & \textbf{NoT (J)} \\
\midrule
GSM8K & GPT-4o-mini & \textbf{89.5} & 69.5 & 82.0 & 85.0 \\
 & Llama-3.3-70B & 91.0 & 76.5 & \textbf{91.5} & \textbf{91.5} \\
 & Qwen2.5-72B & 82.0 & 63.5 & \textbf{91.5} & \textbf{91.5} \\
\addlinespace
Game of 24 & GPT-4o-mini & 58.0 & 11.0 & 75.0 & \textbf{86.0} \\
 & Llama-3.3-70B & 48.0 & 59.0 & 61.0 & \textbf{66.0} \\
 & Qwen2.5-72B & \textbf{48.0} & 38.0 & 38.0 & 42.0 \\
\addlinespace
HotpotQA & GPT-4o-mini & 86.3 & 72.3 & 73.3 & \textbf{91.0} \\
 & Llama-3.3-70B & 68.0 & 73.0 & 73.3 & \textbf{89.0} \\
 & Qwen2.5-72B & 72.3 & 73.7 & 77.3 & \textbf{91.7} \\
\addlinespace
ProofWriter & GPT-4o-mini & 43.3 & \textbf{51.7} & 49.0 & 50.3 \\
 & Llama-3.3-70B & \textbf{79.7} & 61.7 & 57.7 & 57.7 \\
 & Qwen2.5-72B & \textbf{76.3} & 70.7 & 65.0 & 65.0 \\
\bottomrule
\end{tabular}
\end{table}

\begin{table}[t]
\centering
\caption{Cross-model controller comparison (\%). Controller experiments focus on HotpotQA and ProofWriter, the two network-favoring datasets (Table~\ref{tab:datasets}) that also bracket the range of NoT effectiveness: HotpotQA is where NoT shows its largest gain over baselines, and ProofWriter its smallest. HotpotQA evaluated by LLM-as-Judge; ProofWriter by string match. Bold indicates best per dataset-model combination. Weight ablation results on ProofWriter shown below.}
\label{tab:cross_model_rq2}
\small
\begin{tabular}{@{}ll ccc@{}}
\toprule
\textbf{Dataset} & \textbf{Controller} & \textbf{GPT-4o-mini} & \textbf{Llama-3.3-70B} & \textbf{Qwen2.5-72B} \\
\midrule
HotpotQA & Random & 88.7 & 89.0 & 90.3 \\
 & Fixed & 88.0 & 89.3 & \textbf{91.0} \\
 & Self-Gen. & 88.0 & \textbf{89.7} & 90.3 \\
\addlinespace
ProofWriter & Random & 49.7 & \textbf{63.3} & 66.3 \\
 & Fixed & 51.3 & 52.7 & \textbf{69.0} \\
 & Self-Gen. & \textbf{54.0} & 53.7 & 68.7 \\
\midrule
\multicolumn{5}{l}{\textit{Weight ablation (ProofWriter, fixed controller)}} \\
\midrule
 & $w_u$ only & \textbf{57.0} & 58.3 & 58.7 \\
 & $w_d$ only & 51.0 & 58.3 & \textbf{68.0} \\
 & $w_c$ only & 49.0 & 52.3 & 64.7 \\
 & $w_u + w_d$ & 52.3 & \textbf{58.7} & 66.0 \\
 & Full & 49.7 & 54.7 & 66.3 \\
\bottomrule
\end{tabular}
\end{table}

Table~\ref{tab:cross_model} presents the cross-model accuracy comparison and Table~\ref{tab:cross_model_rq2} shows the controller comparison.

\paragraph{NoT benefits scale with model size on sequential tasks.} On GSM8K, both 72B models achieve 91.5\% accuracy with NoT, \emph{outperforming} CoT, the reverse of the GPT-4o-mini pattern where CoT was best (89.5\%). This suggests that larger models can better exploit graph-structured reasoning even on sequential arithmetic, likely because they handle the multi-step graph expansion with fewer compounding errors.

\paragraph{Multi-hop reasoning benefits are consistent.} On HotpotQA, NoT achieves the highest Judge accuracy across all three models: 91.0\% (GPT-4o-mini), 89.0\% (Llama), and 91.7\% (Qwen). The SM-Judge gap remains large and consistent across models (14--18 percentage points), confirming that the evaluation methodology finding generalizes beyond a single model.

\paragraph{CoT dominance on logical reasoning strengthens with model scale.} On ProofWriter, CoT is the best method for both 72B models (79.7\% Llama, 76.3\% Qwen), consistent with the GPT-4o-mini finding. However, the absolute accuracy is dramatically higher: CoT jumps from 43.3\% (GPT-4o-mini) to 79.7\% (Llama), a 36.4 percentage point improvement, suggesting that GPT-4o-mini's low ProofWriter scores partly reflect model capability rather than topology limitations. The topology-task matching principle, that sequential tasks favor chain structure, holds across all three models.

\paragraph{Controller strategy effects are model-dependent.} On HotpotQA, controller choice has minimal impact across all models (88--91\% Judge accuracy). On ProofWriter, the optimal controller varies: self-generated is best for GPT-4o-mini (54.0\%), random for Llama (63.3\%), and fixed for Qwen (69.0\%). The weight ablation reveals model-specific patterns: uncertainty-only ($w_u$) is best for GPT-4o-mini (57.0\%), while dependency-only ($w_d$) is best for Qwen (68.0\%), suggesting that different models benefit from different node prioritization signals.

\section{Further Experiments}
\label{sec:further}

We conduct three additional experiments to probe the boundaries of NoT reasoning: hard mathematical problems that test topology emergence at scale, multi-agent parallelization that tests speedup via concurrent node expansion, and small open-source models (1.5B--7B) that test the minimum capability needed for graph reasoning. All further experiments use GPT-4o-mini unless otherwise noted.

\paragraph{Hard mathematical reasoning.}

We apply CoT, ToT, and NoT to three problems beyond the benchmark suite: IMO~2024 Problem~6 (functional equation), IMO~2023 Problem~6 (geometric proof), and an exploratory decomposition of the Collatz Conjecture into seven sub-tasks. None of the methods produce correct solutions for the IMO problems, consistent with the known difficulty of competition mathematics for current LLMs. However, NoT's reasoning graphs reveal emergent structural properties absent from the linear (CoT) and tree (ToT) topologies.

\begin{figure}[t]
\centering
\includegraphics[width=\textwidth]{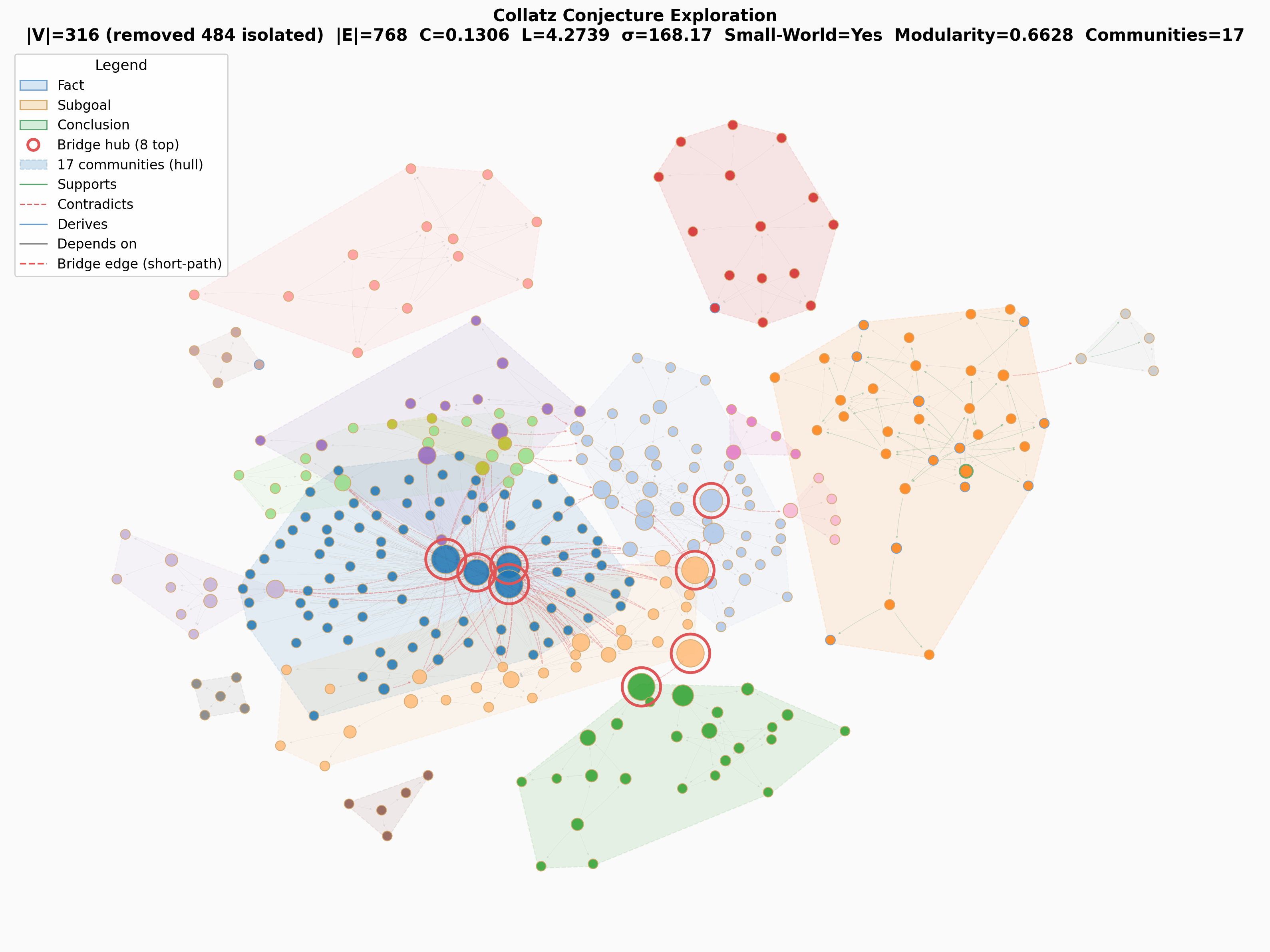}
\caption{NoT reasoning graph for the Collatz Conjecture exploration (800 nodes, 768 edges). Isolated nodes are removed for clarity; 316 connected nodes form 17 communities (colored hulls). Bridge hubs (circled) connect reasoning clusters. Node colors indicate type: blue (fact), orange (subgoal), green (conclusion). Edge types shown in legend.}
\label{fig:collatz_graph}
\end{figure}

\begin{table}[t]
\centering
\caption{Hard problem analysis. NoT generates large-scale graphs with small-world structure ($\sigma \gg 1$). Tokens/node shows NoT reuses sub-conclusions more efficiently than ToT's fixed tree.}
\label{tab:hard_problems}
\small
\begin{tabular}{@{}l rrr rrr r@{}}
\toprule
& \multicolumn{3}{c}{\textbf{Total tokens}} & \multicolumn{4}{c}{\textbf{NoT graph properties}} \\
\cmidrule(lr){2-4} \cmidrule(l){5-8}
\textbf{Problem} & \textbf{CoT} & \textbf{ToT} & \textbf{NoT} & \textbf{Nodes} & \textbf{Edges} & \textbf{$\sigma$} & \textbf{Tok/Node} \\
\midrule
IMO-24 & 1,011 & 334,039 & 66,433 & 130 & 128 & 32.1 & 511 \\
IMO-23 & 1,485 & 365,190 & 111,301 & 151 & 141 & 40.8 & 737 \\
Collatz & 1,490 & 243,977 & 645,104 & 800 & 768 & 168.2 & 806 \\
\bottomrule
\end{tabular}
\end{table}

\begin{figure}[t]
\centering
\includegraphics[width=\textwidth]{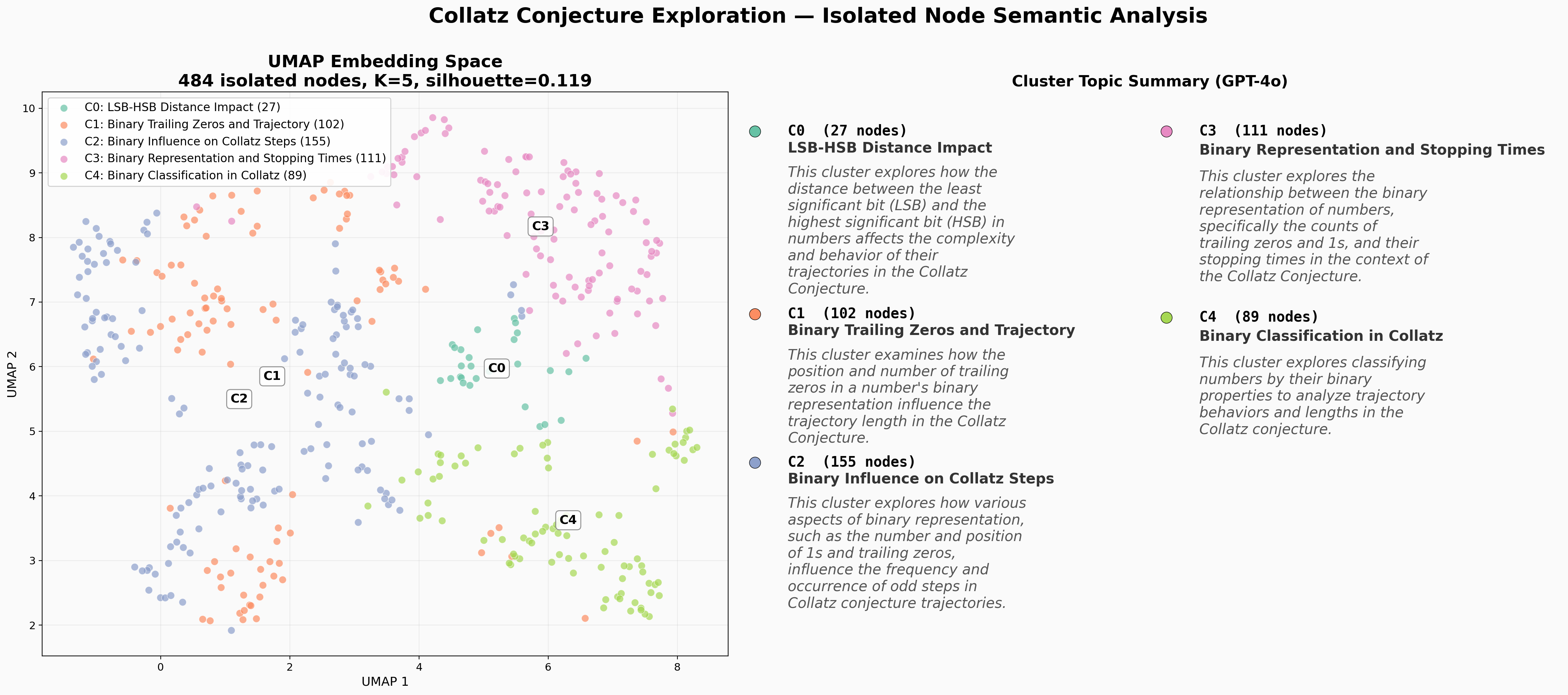}
\caption{Semantic analysis of 484 isolated nodes from the Collatz reasoning graph. Left: UMAP embedding with $K{=}5$ clusters (silhouette $= 0.119$). Right: GPT-4o-generated topic summaries for each cluster, revealing five distinct reasoning themes: LSB--MSB distance impact, binary trailing zeros and trajectory, binary influence on Collatz steps, binary representation and stopping times, and binary classification.}
\label{fig:collatz_clusters}
\end{figure}

Table~\ref{tab:hard_problems} reports token usage and graph properties. NoT generates graphs with 130--800 nodes, far larger than the benchmark tasks (4.9--19.5 nodes in Table~\ref{tab:rq3}), demonstrating that the framework scales to complex problem structures. The NoT graphs exhibit strong small-world topology: clustering coefficients 15--77$\times$ higher than size-matched random networks, yielding small-world coefficients $\sigma = 32$--$168$ ($\sigma > 1$ indicates small-world structure). This contrasts with ToT's tree, which has zero clustering by definition. The degree distributions follow heavy-tailed patterns with hub nodes (degree up to 93 on Collatz), resembling scale-free networks. Figure~\ref{fig:hard_problems} (Appendix~\ref{app:hard_graphs}) illustrates these properties. Figure~\ref{fig:collatz_graph} visualizes the full Collatz reasoning graph (800 nodes, 316 connected after removing 484 isolated nodes), which exhibits clear community structure (17 communities, modularity $Q = 0.66$) with bridge hub nodes connecting reasoning clusters. Figure~\ref{fig:collatz_clusters} shows a UMAP embedding of the 484 isolated nodes, clustered into five semantically coherent topics (e.g., LSB--MSB distance impact, binary trailing zeros, stopping times) identified by GPT-4o. The IMO problem graphs and their isolated-node analyses are provided in Appendix~\ref{app:hard_graphs}.

\paragraph{Multi-agent parallelization.}

Standard NoT expands one node per controller step. We extend the framework to expand $K$ nodes concurrently, where $K$ independent LLM calls process different unresolved nodes in parallel and their results are merged into the shared graph. We evaluate $K \in \{1, 2, 3, 4\}$ on 100 HotpotQA instances.

\begin{table}[t]
\centering
\caption{Multi-agent NoT results on HotpotQA (100 instances). Parallel expansion achieves up to 1.65$\times$ speedup with minimal accuracy or token overhead. Graph topology remains stable across all $K$.}
\label{tab:multi_agent}
\small
\begin{tabular}{@{}c cc rr cc@{}}
\toprule
\textbf{$K$} & \textbf{SM (\%)} & \textbf{Judge (\%)} & \textbf{Avg Tok} & \textbf{Avg Time (s)} & \textbf{Speedup} & \textbf{Tok Overhead} \\
\midrule
1 & 74.0 & 90.0 & 15,229 & 27.2 & 1.00$\times$ & 1.00$\times$ \\
2 & 73.0 & 87.0 & 15,345 & 16.7 & 1.63$\times$ & 1.01$\times$ \\
3 & 71.0 & 88.0 & 16,375 & 19.3 & 1.41$\times$ & 1.08$\times$ \\
4 & 73.0 & 87.0 & 15,923 & 16.5 & \textbf{1.65$\times$} & 1.05$\times$ \\
\bottomrule
\end{tabular}
\end{table}

Table~\ref{tab:multi_agent} reports the results. $K{=}4$ achieves 1.65$\times$ wall-clock speedup over sequential expansion with only 1.05$\times$ token overhead. Semantic accuracy (Judge) remains stable at 87--90\% across all $K$; the slight string-match variation (71--74\%) reflects surface-form differences rather than reasoning degradation. Graph topology is also stable: average node count (5.1--5.3) and reuse rate (0.226--0.238) show no significant change with parallelism, confirming that concurrent expansion does not distort the reasoning structure. Figure~\ref{fig:multi_agent} (Appendix~\ref{app:hard_graphs}) visualizes the accuracy--speedup trade-off.

\paragraph{Small model generalization.}

To determine the minimum model capability for graph reasoning, we run NoT on HotpotQA (100 instances) with five open-source models from 1.5B to 7B parameters, served locally via vLLM at zero API cost. We also include Llama-3.3-70B for comparison.

\begin{table}[t]
\centering
\caption{NoT with small open-source models on HotpotQA (100 instances). Even 1.5B-parameter models execute the NoT pipeline. Graph complexity (nodes, reuse rate) scales with model capability.}
\label{tab:small_models}
\small
\begin{tabular}{@{}llccrcc@{}}
\toprule
\textbf{Model} & \textbf{Params} & \textbf{SM (\%)} & \textbf{Judge (\%)} & \textbf{Avg Tok} & \textbf{Avg Nodes} & \textbf{Reuse} \\
\midrule
Qwen2.5-1.5B & 1.5B & 54.0 & 64.0 & 9,734 & 2.6 & 0.036 \\
Qwen2.5-3B & 3.0B & 58.0 & 69.0 & 18,032 & 5.4 & 0.111 \\
Llama-3.2-3B & 3.0B & 61.0 & 73.0 & 25,390 & 11.2 & 0.173 \\
Phi-3.5-mini & 3.8B & \textbf{64.0} & \textbf{78.0} & 18,648 & 8.1 & 0.187 \\
Qwen2.5-7B & 7.0B & \textbf{64.0} & 77.0 & 14,991 & 4.6 & 0.244 \\
\addlinespace
GPT-4o-mini & --- & 74.0 & 90.0 & 15,229 & 5.2 & 0.226 \\
\bottomrule
\end{tabular}
\end{table}

\begin{figure}[t]
\centering
\includegraphics[width=\textwidth]{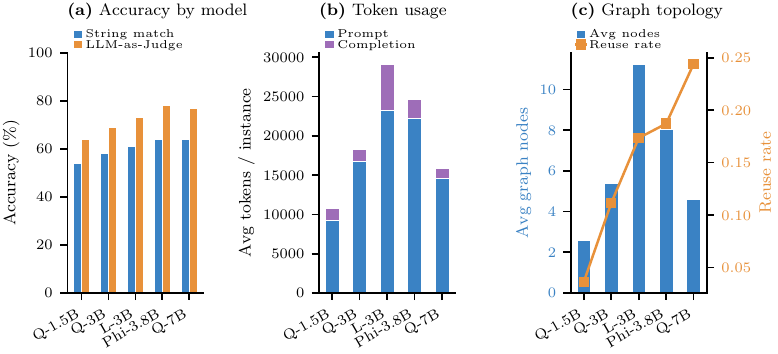}
\caption{NoT with small models on HotpotQA. (a)~Accuracy scales from 64\% (1.5B) to 78\% (3.8B) under Judge evaluation. (b)~Token usage varies by model; Llama-3.2-3B generates the most verbose graphs. (c)~Graph complexity (nodes) and reuse rate increase with model capability, though the relationship is not monotonic.}
\label{fig:small_models}
\end{figure}

Table~\ref{tab:small_models} reports the results. Even Qwen2.5-1.5B executes the full NoT pipeline, achieving 64\% Judge accuracy. The best small model, Phi-3.5-mini (3.8B), reaches 78\% Judge, within 12 percentage points of GPT-4o-mini (90\%). Graph complexity scales with model capability: the \emph{reuse rate} (the fraction of graph nodes with in-degree ${>}\,1$, i.e., nodes that serve as shared sub-conclusions for multiple reasoning paths) increases from 3.6\% (1.5B) to 24.4\% (7B), indicating that larger models build richer reasoning structures that merge evidence rather than duplicating it. Figure~\ref{fig:small_models} shows the scaling trends.

\section{Analysis and Discussion}
\label{sec:discussion}

\subsection{Impact of evaluation methodology}

A striking finding across our experiments is the substantial impact of evaluation methodology on comparative conclusions. On HotpotQA, string-match evaluation places CoT (86.3\%) well ahead of NoT (73.3\%), suggesting a 13.0 percentage point gap. However, LLM-as-Judge evaluation reveals CoT at 95.7\% and NoT at 91.0\%, narrowing the gap to 4.7 percentage points. The string-match metric underestimates NoT by 17.7 percentage points, the largest gap among all methods, because NoT's extraction produces more verbose or reformulated answers that are semantically correct but fail strict string matching.

This finding has broader implications for reasoning research. Comparisons between methods that produce differently formatted outputs may yield misleading conclusions under strict string-match evaluation. We recommend that future work on graph-structured reasoning adopt semantic evaluation to avoid penalizing methods for surface-level format differences.

\subsection{Graph taxonomy emergence}

The NoT framework specifies node and edge types through task-specific prompts but does not hard-constrain the LLM to these types. Across all 900 experimental instances, four node types (\texttt{fact}, \texttt{subgoal}, \texttt{constraint}, \texttt{conclusion}) and four edge types (\texttt{depends\_on}, \texttt{supports}, \texttt{derives}, \texttt{contradicts}) account for 99.8\% of all generated nodes and edges. The most common drift type is \texttt{calculation} (1.4\% of nodes), which emerges in math-heavy tasks as a more specific variant of \texttt{subgoal}. This high adherence to prompted types suggests that LLMs can reliably follow structural schemas for reasoning graph construction.

\subsection{Reasoning graph visualization}

Figure~\ref{fig:not_graphs} shows representative reasoning graphs constructed by NoT for each dataset. The graph structures naturally reflect task characteristics: GSM8K and Game of 24 produce compact, chain-like graphs, while ProofWriter generates larger graphs with many fact and constraint nodes feeding into a single conclusion. HotpotQA produces small but multi-sourced graphs where multiple fact nodes support the conclusion through intermediate subgoals. Detailed per-example analysis with questions, method answers, and node contents is provided in Appendix~\ref{app:examples}.

\subsection{Why NoT underperforms CoT on sequential tasks}

On GSM8K, CoT outperforms NoT by 4.5 percentage points despite NoT's more expressive graph structure. The overhead of graph initialization, multi-step expansion, and answer extraction introduces additional opportunities for error without providing structural benefits for sequential arithmetic. The single-call CoT approach avoids compounding errors across multiple LLM interactions. This confirms that reasoning topology should match task requirements: unnecessary structural complexity degrades rather than improves performance.

\subsection{The premature convergence challenge}

During framework development, we identified that LLMs tend to assign high initial confidence (often 1.0) to all nodes during graph initialization, causing the controller to declare convergence before meaningful reasoning occurs. Our mitigation (clamping initial confidence to 0.5 maximum and enforcing a minimum of 3 expansion steps) partially addresses this, but the tension between premature and delayed convergence remains. On ProofWriter, 70\% of instances still reach the maximum step limit (20 steps) without convergence, indicating that the model struggles to reach high-confidence conclusions on hard logical reasoning problems.

\section{Conclusion}
\label{sec:conclusion}

We proposed Network-of-Thought, a reasoning framework that models LLM reasoning as a directed graph with typed nodes and edges, guided by a heuristic-based controller policy. Our reasoning topology taxonomy formalizes the structural properties and tradeoffs of chain, tree, and network reasoning.

Our experiments across four benchmarks and three model families reveal a nuanced picture along three evaluation dimensions.
\textbf{Accuracy}: CoT remains highly effective for sequential tasks with GPT-4o-mini (89.5\% on GSM8K), while NoT
surpasses ToT on multi-hop reasoning (91.0\% vs.\ 88.0\% on HotpotQA under semantic evaluation). With 72B open-source models, NoT achieves the highest GSM8K accuracy (91.5\%), and Qwen2.5-72B achieves the best multi-hop QA result overall (91.7\% on HotpotQA).
\textbf{Topology simplicity}: NoT produces compact reasoning graphs (4.9 nodes on HotpotQA) that
capture multi-source dependencies without the exponential branching of ToT's BFS tree (21 nodes).
\textbf{Token efficiency}: NoT achieves substantially better token efficiency than ToT on HotpotQA
(16,763 vs.\ 61,270 tokens per correct answer), using moderate computation to deliver competitive
accuracy. Self-generated heuristics further improve efficiency, outperforming fixed and random
strategies on ProofWriter by 2.7 and 4.3 percentage points respectively, with uncertainty-only
scoring achieving the best result at 57.0\%.

Our work also highlights the importance of evaluation methodology. String-match evaluation underestimates reasoning quality by 10 to 20 percentage points on open-ended QA tasks, with the effect being most pronounced for methods that produce structurally different output formats. We recommend that future work on structured reasoning adopt semantic evaluation to ensure fair comparisons.

The key lesson from our investigation is that reasoning topology matters, but not in the way originally hypothesized. Rather than network topology being universally superior for complex tasks, the optimal topology depends on the match between structural requirements and task characteristics. Simple sequential reasoning benefits most from simple chain structure, while multi-source dependency reasoning benefits from graph structure. Cross-model evaluation with Llama-3.3-70B and Qwen2.5-72B confirms that this topology-task matching principle holds across model families, while also revealing that larger models can better exploit graph-structured reasoning: NoT achieves the highest accuracy on GSM8K with 72B models, reversing the GPT-4o-mini pattern. This topology-task matching principle, modulated by model capability, should guide the selection of reasoning frameworks in practice.

\section{Limitations and Future Work}
\label{sec:limitations}

Our primary experiments use GPT-4o-mini with temperature 0.1 and seed 42, supplemented by cross-model evaluation with Llama-3.3-70B-Instruct and Qwen2.5-72B-Instruct (Section~\ref{sec:cross_model}). While the three-model comparison demonstrates generalization across model families, all experiments use a single temperature and seed. Future work should evaluate NoT with frontier reasoning models (GPT-4o, Claude 3.5 Sonnet, Gemini 2.5 Pro) and across multiple random seeds to determine whether even stronger models can further exploit graph-structured reasoning and to quantify run-to-run variance.

The current NoT framework does not generate cyclic reasoning structures. Across all 900 instances, the loop frequency is zero. The LLM does not spontaneously produce hypothesis revision cycles despite prompting that allows them. As reasoning tasks grow more complex across domains such as scientific reasoning, legal argumentation, and medical diagnosis, cyclic structures become increasingly important for iterative hypothesis refinement and evidence reconciliation. Explicit loop-encouraging prompting strategies, or architectural modifications that force iterative refinement, are needed to realize the theoretical advantage of network topology for such tasks.

The evaluation with LLM-as-Judge introduces a dependency on the judge model's quality. While we use the same model (GPT-4o-mini) as the judge for consistency and apply it to all methods for fairness, a stronger or independent judge model would provide more robust evaluation. The judge is also applied only to items that fail string match, so its impact varies across methods. Future work should investigate the sensitivity of comparative conclusions to judge model choice.

Finally, our evaluation covers four benchmarks spanning arithmetic, combinatorial search, multi-hop QA, and logical deduction, but many complex reasoning domains remain unexplored. Extending NoT to tasks such as scientific reasoning, legal argumentation, medical diagnosis, and multi-step planning would test whether graph-structured reasoning generalizes to domains with richer dependency structures, larger evidence spaces, and domain-specific constraint types. Scaling to problems that require dozens or hundreds of reasoning nodes, rather than the 5--21 nodes observed in our experiments, would further stress-test the controller's prioritization strategy and the framework's convergence properties. Our cross-model results suggest that 72B models can better exploit graph-structured reasoning than GPT-4o-mini (e.g., NoT achieves 91.5\% on GSM8K with both Llama and Qwen vs.\ 85.0\% with GPT-4o-mini), but investigating whether this trend extends to smaller models (7B--13B) would clarify the minimum model capability needed for effective graph reasoning. Mapping network topology to multi-agent architectures where graph clusters correspond to specialized agent modules is an additional promising direction.

\section*{Acknowledgements}

We thank the authors and institutions who open-sourced the datasets used in this study, including GSM8K~\citep{cobbe2021gsm8k}, HotpotQA~\citep{yang2018hotpotqa}, and ProofWriter~\citep{tafjord2021proofwriter}, as well as the developers of the open-source large language models that enabled our experiments.


{
\small
\bibliographystyle{plainnat}
\bibliography{references}
}


\clearpage
\appendix

\section{Dataset Examples}
\label{app:data_examples}

We present representative examples from each benchmark dataset to illustrate the reasoning challenges that motivate the choice of reasoning topology.

\subsection{GSM8K}
Grade-school math word problems requiring sequential arithmetic. Example: ``Janet's ducks lay 16 eggs per day. She eats three for breakfast every morning and bakes muffins for her friends every day with four. She sells the remainder at the farmers' market daily for \$2 per fresh duck egg. How much in dollars does she make every day at the farmers' market?'' (Answer: 18)

\subsection{Game of 24}
Combinatorial search over arithmetic expressions. Example: ``Use the numbers 2, 8, 9, 13 with basic arithmetic operations (+, $-$, $\times$, /) to make~24. Each number must be used exactly once.'' (Answer: $(13 - 9) \times (8 - 2) = 24$)

\subsection{HotpotQA}
Bridge-type multi-hop questions requiring evidence from two documents. Example: ``What screenwriter with credits for `Evolution' co-wrote a film starring Nicolas Cage and T\'{e}a Leoni?'' (Answer: David Weissman)

\subsection{ProofWriter}
Open-world logical deduction (depth-5) with negation. Example: Given a set of facts and rules about animals (the cow, lion, tiger, squirrel), determine whether the hypothesis ``The tiger is green'' is True, False, or Unknown. (Answer: False)

\clearpage
\section{Additional Experimental Results}
\label{app:additional}

\subsection{Token efficiency comparison}

\begin{figure}[ht]
\centering
\includegraphics[width=\textwidth]{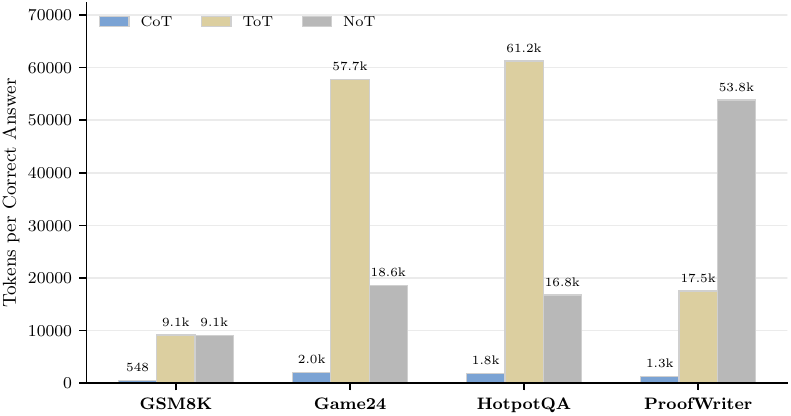}
\caption{Tokens per correct answer by method and dataset. Lower is better. CoT achieves the best token efficiency across all datasets. NoT is substantially more efficient than ToT on HotpotQA.}
\label{fig:token_efficiency}
\end{figure}

\subsection{String match vs. LLM-as-Judge gap}

\begin{figure}[ht]
\centering
\includegraphics[width=\textwidth]{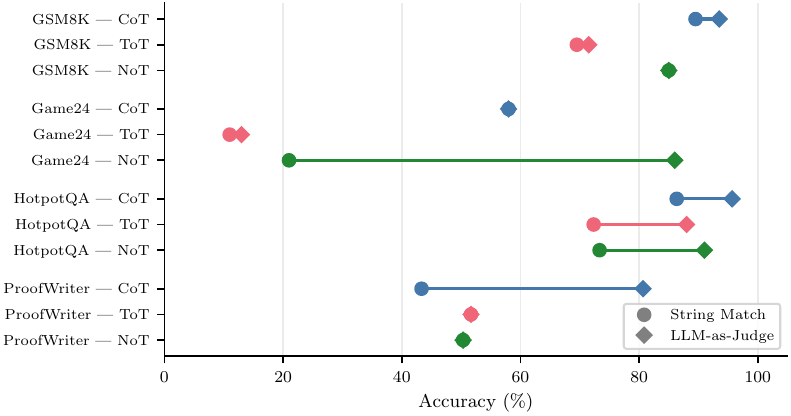}
\caption{Accuracy gap between string match and LLM-as-Judge evaluation across all methods and datasets. NoT exhibits the largest gap on HotpotQA (17.7 percentage points), indicating that its more verbose extraction produces semantically correct but format-mismatched answers.}
\label{fig:judge_gap}
\end{figure}

\subsection{Topology comparison: Rev1 vs. Rev3}

\begin{figure}[ht]
\centering
\includegraphics[width=\textwidth]{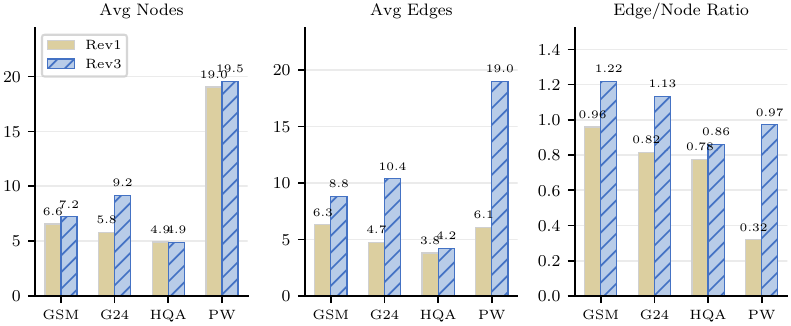}
\caption{Graph topology comparison between NoT Rev1 (before framework fixes) and Rev3 (final version). Rev3 produces denser, better-connected graphs with higher edge-to-node ratios across all datasets.}
\label{fig:topology}
\end{figure}

\subsection{CoT implicit step counting}
\label{app:cot_steps}

To enable topology comparison across all three methods, we operationalize CoT reasoning traces as implicit linear chains by counting the number of reasoning steps within each output. Since CoT produces unstructured text rather than an explicit graph, we apply the following heuristic procedure:

\paragraph{Step detection.} We count numbered list items using the regular expression:

\begin{quote}
\small\ttfamily (?:\string^\,|\,\textbackslash n)\textbackslash s*(?:\textbackslash*\textbackslash*)?(\textbackslash d+)(?:\textbackslash*\textbackslash*)?[.)]\textbackslash s+[A-Z\textbackslash*]
\end{quote}

\noindent This pattern requires five conditions to fire: (1)~the match must begin at a line start or after a newline; (2)~optional leading whitespace is consumed; (3)~optional markdown bold markers (\texttt{**}) surrounding the digit are allowed; (4)~the digit(s) must be followed by a period or closing parenthesis and a space; and (5)~the next character must be an uppercase letter or asterisk (indicating the start of a sentence or bold text). Condition~(1) anchors the match to list items rather than inline references, while condition~(5) excludes numeric values embedded in mathematical expressions such as \texttt{\$16.50} or \texttt{= 117}, where the character after the period is a digit or operator rather than an uppercase letter. The step count for each instance is the maximum numbered item found, which handles non-contiguous numbering and sub-step nesting.

\paragraph{Fallback.} When no numbered steps are detected (e.g., short single-expression answers in Game of~24), we fall back to counting non-empty paragraphs separated by double newlines, with a minimum of~1.

\paragraph{Outlier handling.} A small number of CoT traces contain pathological brute-force enumerations (e.g., 200 numbered attempts on Game of~24). We clip step counts at the 99th percentile per dataset to avoid distorting the averages.

\paragraph{Topology mapping.} Each detected step becomes an implicit node in a linear chain, with edges $=$ nodes $- 1$. This yields average CoT chain lengths of 7.3 (GSM8K), 4.1 (Game of~24), 3.7 (HotpotQA), and 5.1 (ProofWriter) steps. These values are reported in the Nodes column of Table~\ref{tab:rq3}.

\clearpage
\section{Reasoning Graph Examples}
\label{app:examples}

We present four representative NoT reasoning graphs, one per dataset, to illustrate how the framework constructs typed node-edge structures for different reasoning tasks. All methods (CoT, ToT, NoT) answer these examples correctly. Nodes are colored by type: \textcolor[HTML]{6A9ECF}{fact} (blue), \textcolor[HTML]{D4A76A}{subgoal} (orange), \textcolor[HTML]{B07EB0}{constraint} (purple), \textcolor[HTML]{5BA66F}{conclusion} (green). Edge types are \emph{depends\_on} (gray), \emph{supports} (green), \emph{contradicts} (red dashed), and \emph{derives} (blue).

\begin{figure}[ht]
\centering
\includegraphics[width=0.85\textwidth]{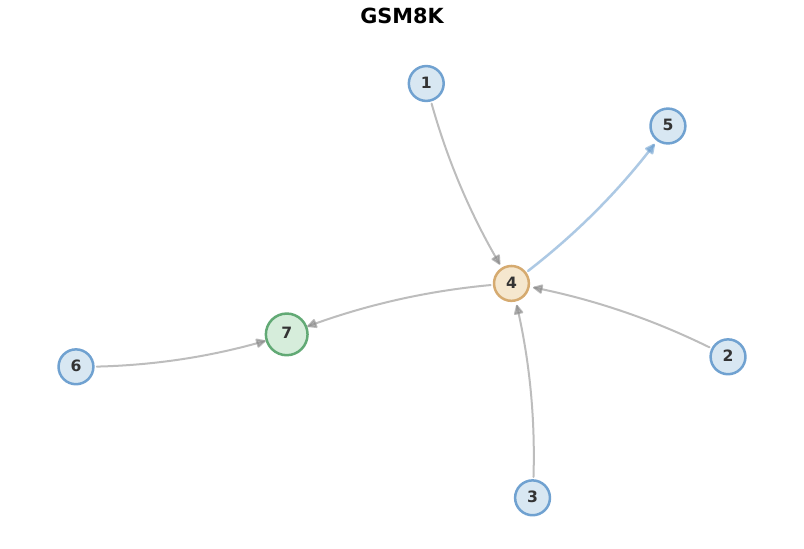}
\caption{GSM8K example (7~nodes, 6~edges). \emph{Question:} ``Janet's ducks lay 16~eggs per day\ldots'' Ground truth: 18. CoT: 18; ToT: \$18; NoT: 18. The graph decomposes the problem into fact nodes (1--3, 5--6) encoding the given quantities, a subgoal node~(4) for computing the number of remaining eggs, and a conclusion node~(7) for the final revenue calculation.}
\label{fig:example_gsm8k}
\end{figure}

\begin{figure}[ht]
\centering
\includegraphics[width=0.85\textwidth]{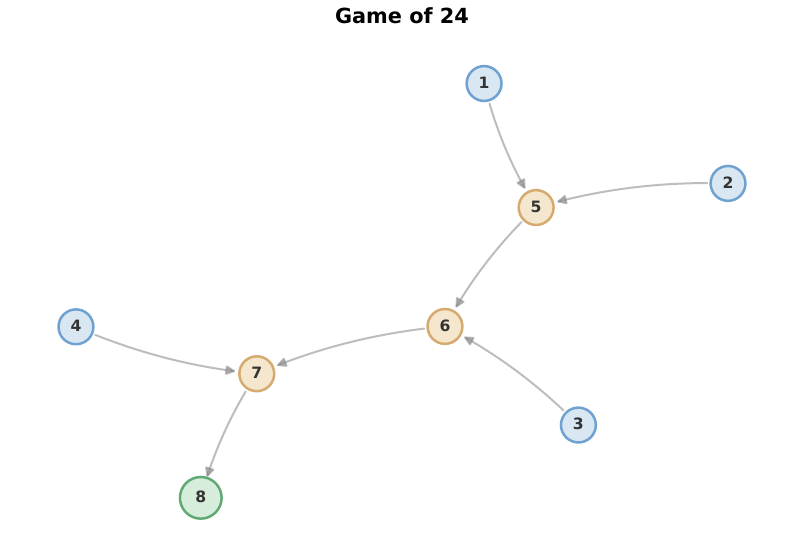}
\caption{Game of 24 example (8~nodes, 7~edges). \emph{Question:} ``Use the numbers 2, 8, 9, 13\ldots to make~24.'' Ground truth: 24. CoT: $(8 \times 2) + (13 - 9)$; ToT: 24; NoT: $(13 - 9) \times (8 - 2) = 24$. The graph encodes each input number as a fact node (1--4), chains arithmetic operations through subgoal nodes (5--7), and produces the conclusion node~(8).}
\label{fig:example_game24}
\end{figure}

\begin{figure}[ht]
\centering
\includegraphics[width=0.85\textwidth]{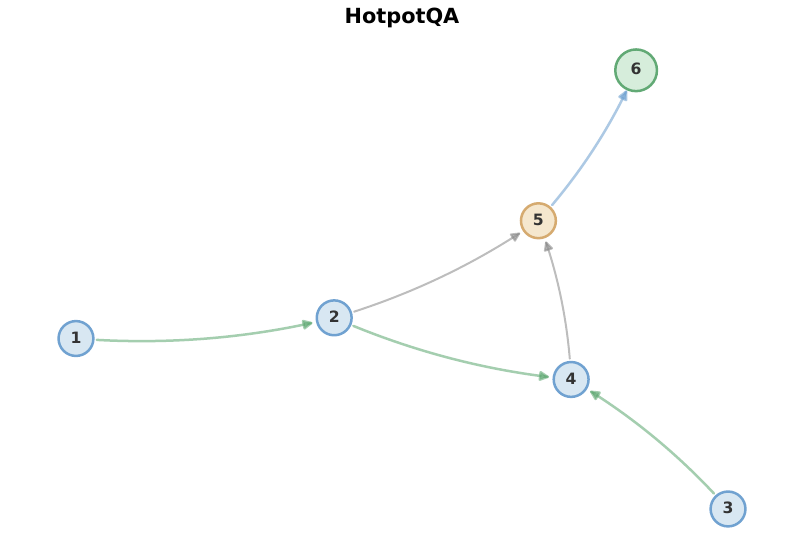}
\caption{HotpotQA example (6~nodes, 6~edges). \emph{Question:} ``What screenwriter with credits for `Evolution' co-wrote a film starring Nicolas Cage and T\'{e}a Leoni?'' Ground truth: David Weissman. The graph gathers evidence from two independent sources (nodes~1--2 and 3--4), converging through a subgoal node~(5) that synthesizes both criteria into the conclusion node~(6). This multi-source convergence pattern is characteristic of bridge-type multi-hop reasoning.}
\label{fig:example_hotpotqa}
\end{figure}

\begin{figure}[ht]
\centering
\includegraphics[width=0.85\textwidth]{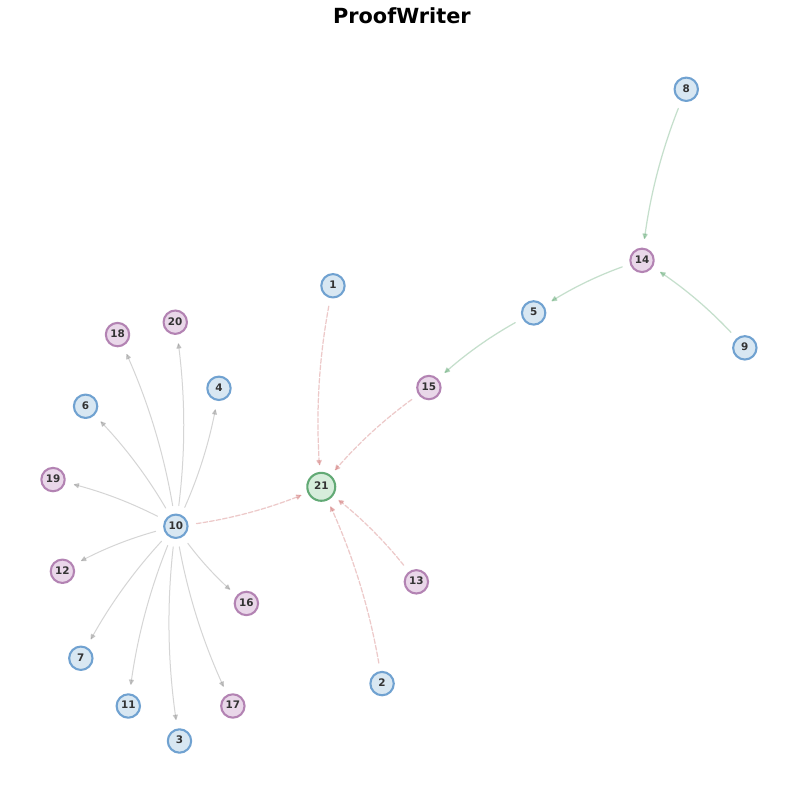}
\caption{ProofWriter example (21~nodes, 20~edges). Fact nodes (1--11) encode given propositions and constraint nodes (12--20) encode conditional rules. The conclusion node~(21) states the hypothesis ``The tiger is green.'' Five \emph{contradicts} edges (dashed red) point into node~21, indicating multiple lines of evidence refuting the hypothesis. Node~10 (``The tiger is not green'') serves as a hub with 11~outgoing \emph{depends\_on} edges.}
\label{fig:example_proofwriter}
\end{figure}

\subsection{Detailed reasoning traces}
\label{app:traces}

We provide the full NoT reasoning graph contents for the four examples visualized in Figures~\ref{fig:example_gsm8k}--\ref{fig:example_proofwriter}. All methods (CoT, ToT, NoT) answer these examples correctly. For each example, we list the question, gold answer, predictions from all three methods, and the complete node and edge contents of the NoT graph.

\subsubsection{GSM8K (index 0, Figure~\ref{fig:example_gsm8k})}

\paragraph{Question.} ``Janet's ducks lay 16 eggs per day. She eats three for breakfast every morning and bakes muffins for her friends every day with four. She sells the remainder at the farmers' market daily for \$2 per fresh duck egg. How much in dollars does she make every day at the farmers' market?''\\
\textbf{Gold:} 18. \textbf{CoT:} 18. \textbf{ToT:} \$18. \textbf{NoT:} 18.

\paragraph{NoT graph (7 nodes, 6 edges).}
\begin{quote}\small
\emph{node\_001} [fact]: Janet's ducks lay 16 eggs per day.\\
\emph{node\_002} [fact]: Janet eats 3 eggs for breakfast every morning.\\
\emph{node\_003} [fact]: Janet bakes muffins with 4 eggs every day.\\
\emph{node\_004} [subgoal]: Calculate the number of eggs Janet sells at the farmers' market.\\
\emph{node\_005} [fact]: Janet sells the remainder of the eggs at the farmers' market.\\
\emph{node\_006} [fact]: Janet sells each egg for \$2.\\
\emph{node\_007} [conclusion]: Calculate the total revenue Janet makes from selling eggs.\\[4pt]
\emph{Edges:} node\_001 $\xrightarrow{\text{depends\_on}}$ node\_004; node\_002 $\xrightarrow{\text{depends\_on}}$ node\_004; node\_003 $\xrightarrow{\text{depends\_on}}$ node\_004; node\_004 $\xrightarrow{\text{derives}}$ node\_005; node\_004 $\xrightarrow{\text{depends\_on}}$ node\_007; node\_006 $\xrightarrow{\text{depends\_on}}$ node\_007.\\[4pt]
\emph{Note: The conclusion node describes the task rather than the numeric answer. The answer 18 is computed by the extraction step from the subgoal chain: $16 - 3 - 4 = 9$ remaining eggs, $9 \times \$2 = \$18$.}
\end{quote}

\subsubsection{Game of 24 (index 16, Figure~\ref{fig:example_game24})}

\paragraph{Question.} ``Use the numbers 2, 8, 9, 13 with basic arithmetic operations (+, -, *, /) to make 24. Each number must be used exactly once.''\\
\textbf{Gold:} 24. \textbf{CoT:} $(8 \times 2) + (13 - 9)$. \textbf{ToT:} 24. \textbf{NoT:} $(13 - 9) \times (8 - 2) = 24$.

\paragraph{NoT graph (8 nodes, 7 edges).}
\begin{quote}\small
\emph{node\_001} [fact]: Number 2.\\
\emph{node\_002} [fact]: Number 8.\\
\emph{node\_003} [fact]: Number 9.\\
\emph{node\_004} [fact]: Number 13.\\
\emph{node\_005} [subgoal]: Combine 8 and 2 to get 10.\\
\emph{node\_006} [subgoal]: Combine 10 and 9 to get 19.\\
\emph{node\_007} [subgoal]: Combine 19 and 13 to get 6.\\
\emph{node\_008} [conclusion]: Combine 6 and 24 to get 24.\\[4pt]
\emph{Edges:} node\_001 $\xrightarrow{\text{depends\_on}}$ node\_005; node\_002 $\xrightarrow{\text{depends\_on}}$ node\_005; node\_005 $\xrightarrow{\text{depends\_on}}$ node\_006; node\_003 $\xrightarrow{\text{depends\_on}}$ node\_006; node\_006 $\xrightarrow{\text{depends\_on}}$ node\_007; node\_004 $\xrightarrow{\text{depends\_on}}$ node\_007; node\_007 $\xrightarrow{\text{depends\_on}}$ node\_008.
\end{quote}

\subsubsection{HotpotQA (index 8, Figure~\ref{fig:example_hotpotqa})}

\paragraph{Question.} ``What screenwriter with credits for `Evolution' co-wrote a film starring Nicolas Cage and T\'{e}a Leoni?''\\
\textbf{Gold:} David Weissman. \textbf{CoT:} David Weissman. \textbf{ToT:} David Weissman. \textbf{NoT:} David Weissman.

\paragraph{NoT graph (6 nodes, 6 edges).}
\begin{quote}\small
\emph{node\_001} [fact]: David Weissman is a screenwriter and director.\\
\emph{node\_002} [fact]: David Weissman has film credits including `Evolution' (2001).\\
\emph{node\_003} [fact]: The Family Man is a film starring Nicolas Cage and T\'{e}a Leoni.\\
\emph{node\_004} [fact]: David Weissman co-wrote The Family Man.\\
\emph{node\_005} [subgoal]: Identify a screenwriter who has credits for `Evolution' and co-wrote a film starring Nicolas Cage and T\'{e}a Leoni.\\
\emph{node\_006} [conclusion]: David Weissman is the screenwriter who co-wrote a film starring Nicolas Cage and T\'{e}a Leoni and has credits for `Evolution'.\\[4pt]
\emph{Edges:} node\_001 $\xrightarrow{\text{supports}}$ node\_002; node\_003 $\xrightarrow{\text{supports}}$ node\_004; node\_002 $\xrightarrow{\text{depends\_on}}$ node\_005; node\_004 $\xrightarrow{\text{depends\_on}}$ node\_005; node\_005 $\xrightarrow{\text{derives}}$ node\_006; node\_002 $\xrightarrow{\text{supports}}$ node\_004.\\[4pt]
\emph{Note: This graph exhibits multi-source convergence: two independent evidence chains (nodes 1--2 and 3--4) converge through the subgoal node (5) into the conclusion (6), characteristic of bridge-type multi-hop reasoning.}
\end{quote}

\subsubsection{ProofWriter (index 1, Figure~\ref{fig:example_proofwriter})}

\paragraph{Question.} Facts: The cow is not big. The cow is not green. The lion eats the tiger. The lion sees the cow. The lion visits the cow. The lion does not visit the squirrel. The lion visits the tiger. The squirrel is big. The squirrel is round. The tiger is not green. The tiger does not see the cow. Rules: If something sees the squirrel and the squirrel eats the cow then the cow is round. If something is green then it eats the tiger. [+7 additional rules.] \textbf{Hypothesis:} The tiger is green.\\
\textbf{Gold:} False. \textbf{CoT:} False. \textbf{ToT:} False. \textbf{NoT:} False.

\paragraph{NoT graph (21 nodes, 20 edges).}
\begin{quote}\small
\emph{Fact nodes (1--11):} The cow is not big; The cow is not green; The lion eats the tiger; The lion sees the cow; The lion visits the cow; The lion does not visit the squirrel; The lion visits the tiger; The squirrel is big; The squirrel is round; The tiger is not green; The tiger does not see the cow.\\[4pt]
\emph{Constraint nodes (12--20):} If something sees the squirrel and the squirrel eats the cow then the cow is round; If something is green then it eats the tiger; If the squirrel is round then the squirrel visits the cow; If something eats the cow then it sees the squirrel; If something sees the tiger and the tiger visits the squirrel then it is nice; If something is round then it eats the cow; If something is kind then it eats the cow; If the tiger visits the cow then the cow sees the squirrel; If something sees the cow then the cow eats the tiger.\\[4pt]
\emph{node\_021} [conclusion]: The tiger is green.\\[4pt]
\emph{Key structural features:} Five \emph{contradicts} edges point into the conclusion node (from nodes 1, 2, 10, 13, 15), indicating multiple lines of evidence refuting the hypothesis. Node 10 (``The tiger is not green'') serves as a hub with 11 outgoing \emph{depends\_on} edges connecting to nodes 3--7, 11--12, and 16--20.
\end{quote}

\clearpage
\section{Hard Problem Reasoning Graphs}
\label{app:hard_graphs}

\begin{figure}[ht]
\centering
\includegraphics[width=\textwidth]{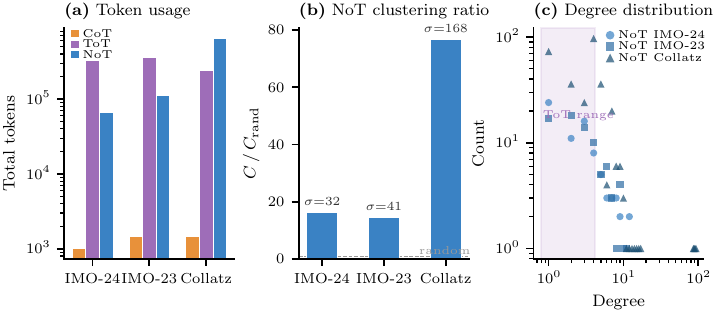}
\caption{Hard problem analysis. (a)~Token usage across methods (log scale). (b)~NoT clustering ratio $C/C_{\text{rand}}$ with small-world coefficient $\sigma$; the dashed line marks random-graph baseline. (c)~NoT degree distributions on log--log axes; the shaded region shows the narrow ToT range (degree 1--4). NoT produces heavy-tailed distributions with high-degree hub nodes.}
\label{fig:hard_problems}
\end{figure}

\begin{figure}[ht]
\centering
\includegraphics[width=\textwidth]{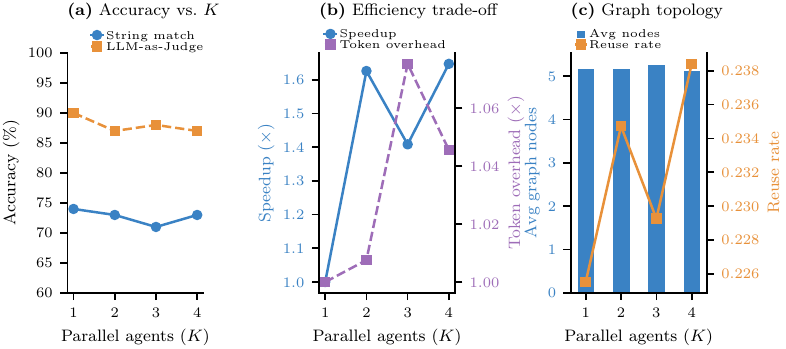}
\caption{Multi-agent NoT on HotpotQA. (a)~Accuracy is stable across $K$ (Judge 87--90\%). (b)~Speedup increases with $K$ while token overhead stays near 1.0$\times$. (c)~Graph topology (nodes and reuse rate) is preserved under parallelism.}
\label{fig:multi_agent}
\end{figure}

This appendix collects supplementary figures from Section~\ref{sec:further}. Figure~\ref{fig:hard_problems} shows the token usage and small-world analysis for the hard problem experiments. Figure~\ref{fig:multi_agent} shows the multi-agent parallelization trade-offs. The remaining subsections provide the full NoT reasoning graph visualizations and isolated-node semantic analyses for the two IMO competition problems. The Collatz Conjecture visualizations are shown in Figures~\ref{fig:collatz_graph}--\ref{fig:collatz_clusters} in the main text.

\paragraph{Graph layout.} Each reasoning graph is visualized using a spring-force layout (Fruchterman--Reingold) applied to the connected component after removing isolated nodes (in-degree $=$ out-degree $= 0$). Nodes are colored by type (blue: fact, orange: subgoal, green: conclusion, red: constraint/hypothesis) and sized proportionally to degree. Community structure is detected via the Louvain algorithm and shown as colored convex hulls. Bridge hub nodes, identified as the top-$k$ nodes by betweenness centrality, are highlighted with circles. Edge types (supports, contradicts, derives, depends\_on) are distinguished by color and style.

\paragraph{Isolated node analysis.} Nodes with no edges represent reasoning units that the LLM generated but never connected to the main graph. To understand what these isolated nodes represent, we embed their textual content using OpenAI's \texttt{text-embedding-3-small} model, project into two dimensions via UMAP, and cluster with $K$-means (selecting $K$ by silhouette score). Each cluster is then summarized by prompting GPT-4o with the concatenated node contents and asking for a short topic label and description. This pipeline reveals what reasoning directions the model explored but failed to integrate into the main graph structure.

\subsection{IMO 2024 Problem 6}

\begin{figure}[ht]
\centering
\includegraphics[width=\textwidth]{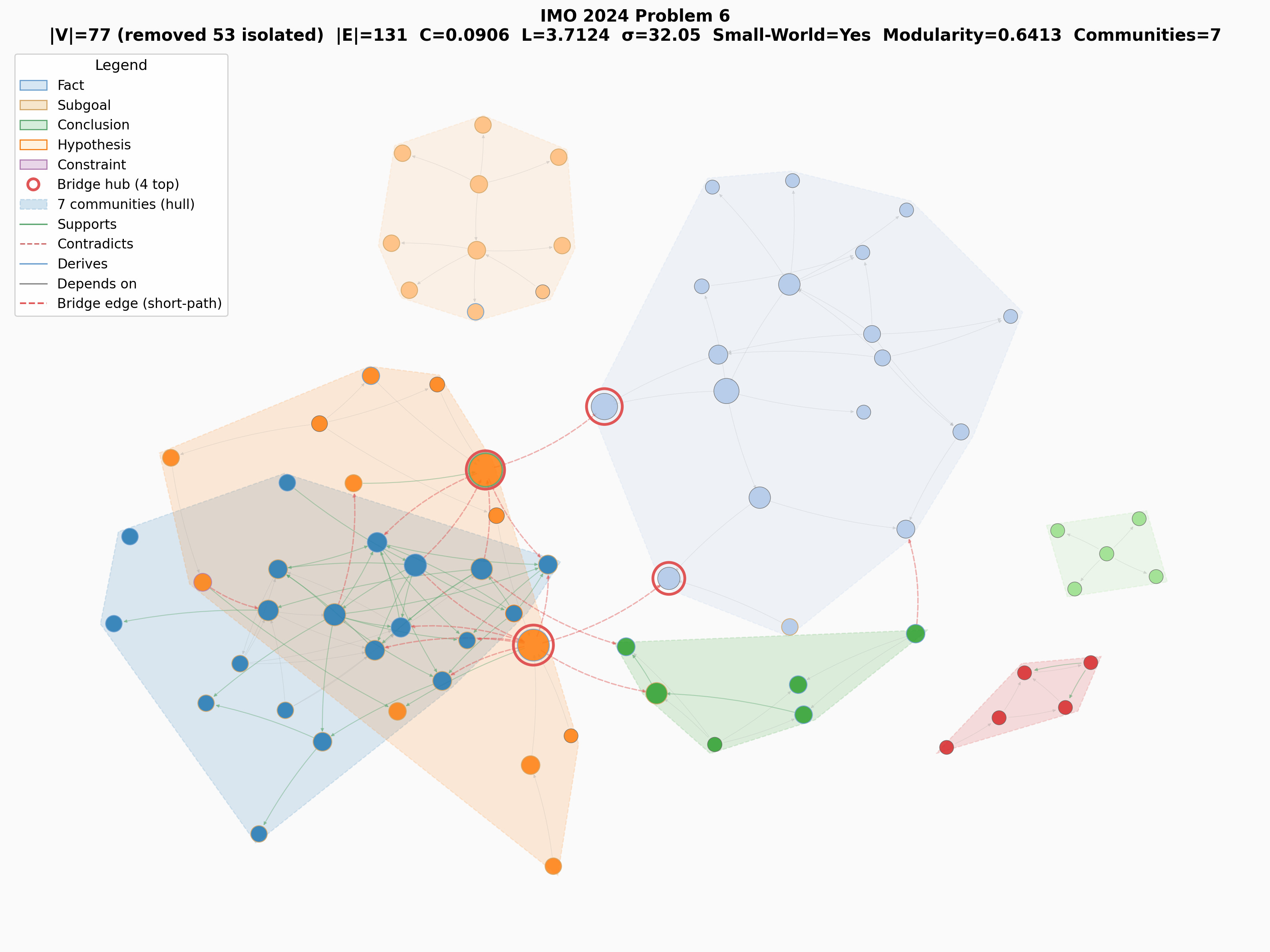}
\caption{NoT reasoning graph for IMO 2024 Problem 6 (functional equation). 77 connected nodes remain after removing 53 isolated nodes. The graph forms 7 communities ($Q = 0.64$, $\sigma = 32.1$). Hub nodes (circled) connect reasoning clusters spanning constraint analysis, injectivity, and surjectivity sub-problems.}
\label{fig:imo2024_graph}
\end{figure}

\begin{figure}[ht]
\centering
\includegraphics[width=\textwidth]{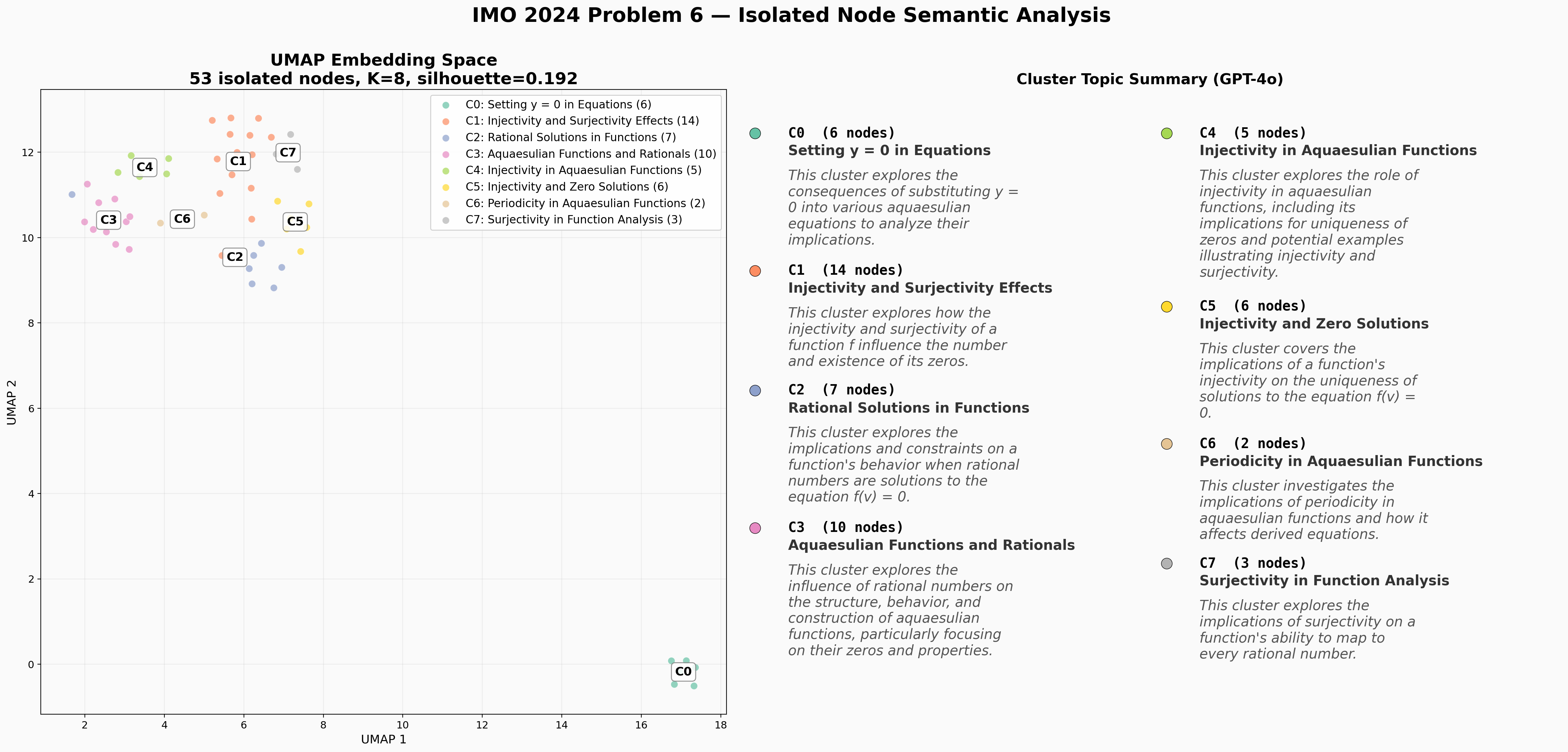}
\caption{Semantic analysis of 53 isolated nodes from IMO 2024 Problem 6. UMAP embedding with $K{=}8$ clusters (silhouette $= 0.192$). GPT-4o identifies distinct reasoning themes including setting $y = 0$ in equations, injectivity/surjectivity effects, rational solutions, and aquasequian function properties.}
\label{fig:imo2024_clusters}
\end{figure}

\subsection{IMO 2023 Problem 6}

\begin{figure}[ht]
\centering
\includegraphics[width=\textwidth]{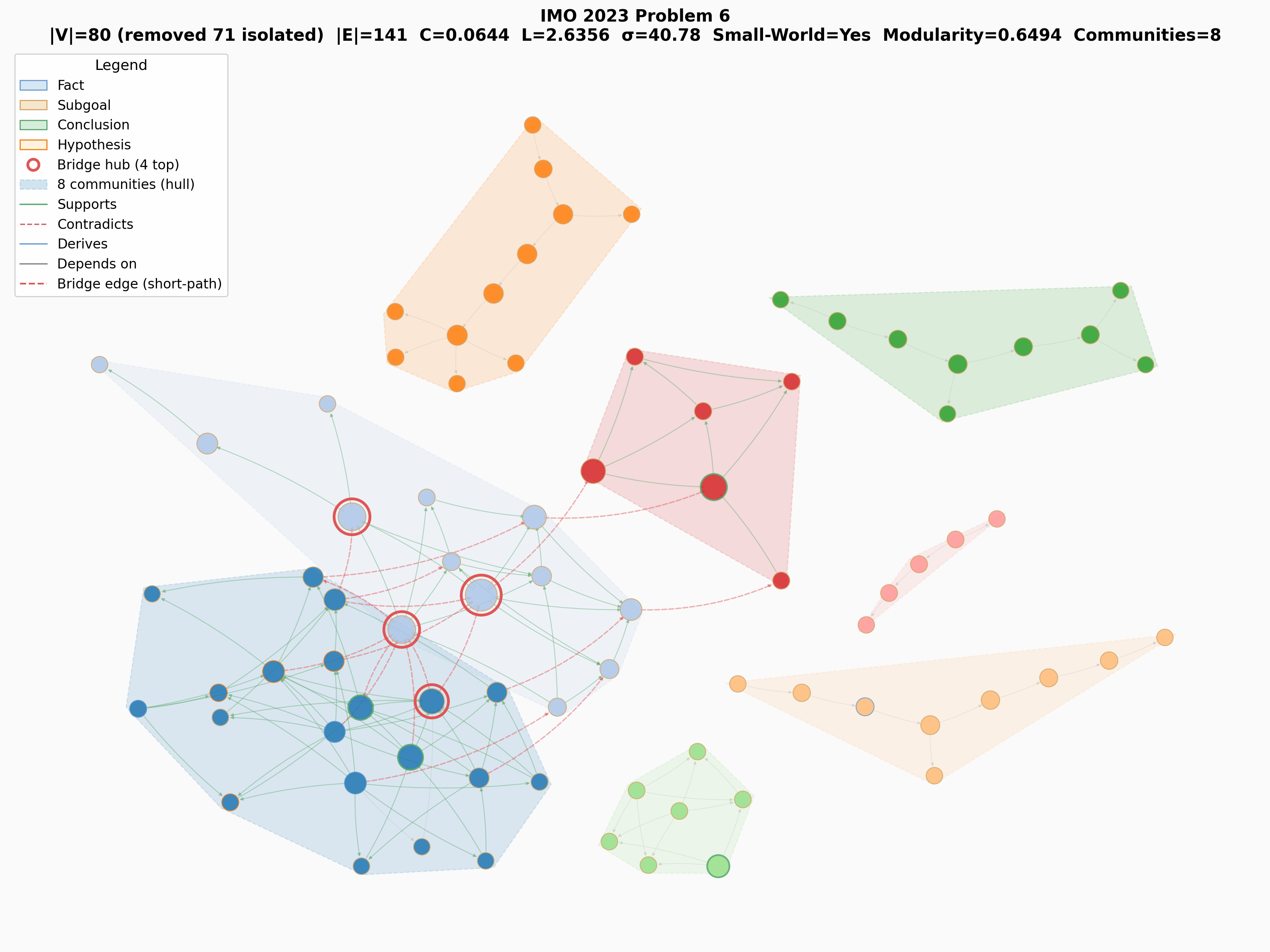}
\caption{NoT reasoning graph for IMO 2023 Problem 6 (geometric proof involving circumcircles of an equilateral triangle). 80 connected nodes remain after removing 71 isolated nodes. The graph forms 8 communities ($Q = 0.65$, $\sigma = 40.8$). Reasoning clusters correspond to angle calculations, distance computations, and geometric verification strategies.}
\label{fig:imo2023_graph}
\end{figure}

\begin{figure}[ht]
\centering
\includegraphics[width=\textwidth]{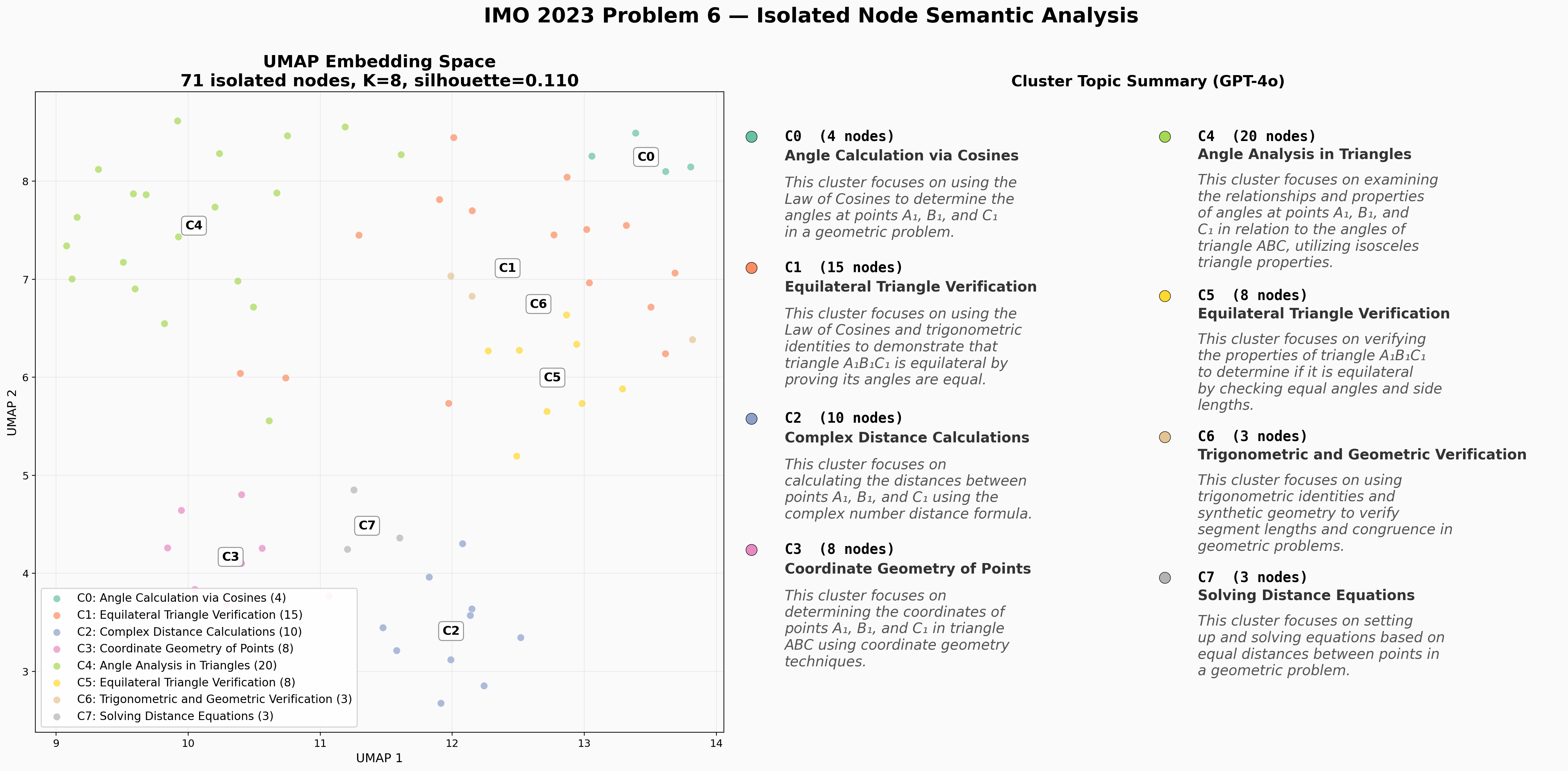}
\caption{Semantic analysis of 71 isolated nodes from IMO 2023 Problem 6. UMAP embedding with $K{=}8$ clusters (silhouette $= 0.110$). GPT-4o identifies reasoning themes including angle calculation via cosines, equilateral triangle verification, complex distance calculations, coordinate geometry, and trigonometric verification.}
\label{fig:imo2023_clusters}
\end{figure}

\clearpage
\section{Task-Specific Prompts}
\label{app:prompts}

We provide the full prompts used for graph initialization (Phase~1) and answer extraction (Phase~6) across all four datasets. Each initialization prompt instructs the LLM to produce a JSON reasoning graph with typed nodes and edges; each extraction prompt anchors on the conclusion node's value to produce the final answer.

\subsection{Graph initialization prompts}

\paragraph{GSM8K.}
\begin{quote}\small\ttfamily
Analyze this math problem and decompose it into reasoning steps.
For each step, output a JSON node with:
- "id": unique string like "node\_001"
- "type": one of "fact" (given info), "subgoal" (intermediate calculation), "conclusion" (final answer)
- "content": description of this step
- "confidence": float 0.0--1.0
- "depends\_on": list of node IDs this depends on

You MUST include exactly one "conclusion" node that represents the final numeric answer.
List edges: "source", "target", "type" (supports/depends\_on/derives), "weight" (0.0--1.0).

Output: \{``nodes'': [...], ``edges'': [...]\}

Problem: \{problem\}
\end{quote}

\paragraph{Game of 24.}
\begin{quote}\small\ttfamily
Analyze this Game of 24 problem. You must find an arithmetic expression using +, -, *, / that equals 24 using all four given numbers exactly once.
Decompose into nodes:
- "id": unique string like "node\_001"
- "type": one of "fact" (input numbers), "subgoal" (intermediate combination), "conclusion" (final expression)
- "content": description
- "confidence": float 0.0--1.0
- "depends\_on": list of node IDs

You MUST include exactly one "conclusion" node for the final expression.
List edges: "source", "target", "type" (supports/depends\_on/derives), "weight".

Output: \{``nodes'': [...], ``edges'': [...]\}

Problem: \{problem\}
\end{quote}

\paragraph{HotpotQA.}
\begin{quote}\small\ttfamily
Analyze this multi-hop question. Identify facts from the context and reasoning steps needed.
For each component, output a JSON node:
- "id": unique string like "node\_001"
- "type": one of "fact" (from context), "subgoal" (intermediate reasoning), "conclusion" (final answer)
- "content": description
- "confidence": float 0.0--1.0
- "depends\_on": list of node IDs

You MUST include exactly one "conclusion" node for the final answer.
List edges: "source", "target", "type" (supports/depends\_on/derives), "weight".

Output: \{``nodes'': [...], ``edges'': [...]\}

Problem: \{problem\}
\end{quote}

\paragraph{ProofWriter.}
\begin{quote}\small\ttfamily
Analyze this logical reasoning problem. Identify facts, rules, and the conclusion to prove.
For each component, output a JSON node:
- "id": unique string like "node\_001"
- "type": one of "fact" (given statement), "constraint" (rule/implication), "subgoal" (intermediate deduction), "conclusion" (the statement to prove/disprove)
- "content": description
- "confidence": float 0.0--1.0
- "depends\_on": list of node IDs

You MUST include exactly one "conclusion" node representing the statement to be proved True, False, or Unknown.
List edges: "source", "target", "type" (supports/contradicts/depends\_on/derives), "weight".

Output: \{``nodes'': [...], ``edges'': [...]\}

Problem: \{problem\}
\end{quote}

\subsection{Answer extraction prompts}

\paragraph{GSM8K.}
\begin{quote}\small\ttfamily
Based on the reasoning below, what is the final numeric answer?

Problem: \{problem\}

Reasoning: \{reasoning\_chain\}

The reasoning concluded with the value: \{conclusion\_value\}

Provide ONLY the final number (e.g., 42 or 3.5). Nothing else.
\end{quote}

\paragraph{Game of 24.}
\begin{quote}\small\ttfamily
Based on the reasoning below, write an arithmetic expression that equals exactly 24.

Problem: \{problem\}

Reasoning: \{reasoning\_chain\}

The reasoning concluded with: \{conclusion\_value\}

Your expression MUST:
1. Use all four input numbers exactly once
2. Use only +, -, *, / and parentheses
3. Evaluate to exactly 24

Verify your expression before answering. Provide ONLY the expression. Nothing else.
\end{quote}

\paragraph{HotpotQA.}
\begin{quote}\small\ttfamily
Based on the reasoning below, answer the question.

Problem: \{problem\}

Reasoning: \{reasoning\_chain\}

The reasoning concluded with the answer: \{conclusion\_value\}

Provide the final answer based on the reasoning chain and its conclusion.
Give the complete answer without abbreviation or truncation.
Provide ONLY the answer, nothing else.
\end{quote}

\paragraph{ProofWriter.}
\begin{quote}\small\ttfamily
Based on the reasoning below, determine if the statement is True, False, or Unknown.

Problem: \{problem\}

Reasoning: \{reasoning\_chain\}

The reasoning concluded: \{conclusion\_value\}

Based on the full reasoning chain:
- If the facts and rules conclusively prove it, answer True.
- If the facts and rules conclusively disprove it, answer False.
- If there is insufficient evidence, answer Unknown.

Provide ONLY one word: True, False, or Unknown.
\end{quote}


\clearpage
\section*{NeurIPS Paper Checklist}

\begin{enumerate}

\item {\bf Claims}
    \item[] Question: Do the main claims made in the abstract and introduction accurately reflect the paper's contributions and scope?
    \item[] Answer: \answerYes{}
    \item[] Justification: All accuracy numbers in the abstract are directly drawn from our experimental results (Tables~\ref{tab:main_results},~\ref{tab:rq2},~\ref{tab:rq3}). The claims about topology-task matching are supported by the empirical analysis in Section~\ref{sec:results}.

\item {\bf Limitations}
    \item[] Question: Does the paper discuss the limitations of the work performed by the authors?
    \item[] Answer: \answerYes{}
    \item[] Justification: Section~\ref{sec:limitations} discusses single-model evaluation, small sample size for Game of 24, absence of cyclic structures, LLM-as-Judge limitations, and missing small model experiments.

\item {\bf Theory assumptions and proofs}
    \item[] Question: For each theoretical result, does the paper provide the full set of assumptions and a complete (and correct) proof?
    \item[] Answer: \answerNA{}
    \item[] Justification: This paper presents an empirical framework and experimental analysis. The reasoning topology taxonomy (Section~\ref{sec:taxonomy}) is a classification framework, not a theoretical result requiring formal proof.

    \item {\bf Experimental result reproducibility}
    \item[] Question: Does the paper fully disclose all the information needed to reproduce the main experimental results of the paper to the extent that it affects the main claims and/or conclusions of the paper (regardless of whether the code and data are provided or not)?
    \item[] Answer: \answerYes{}
    \item[] Justification: Section~\ref{sec:eval_metrics} specifies the model, evaluation metrics, and scoring methodology. The framework architecture and controller scoring function are fully described in Section~\ref{sec:framework}.

\item {\bf Open access to data and code}
    \item[] Question: Does the paper provide open access to the data and code, with sufficient instructions to faithfully reproduce the main experimental results, as described in supplemental material?
    \item[] Answer: \answerYes{}
    \item[] Justification: Code and all result files will be released upon publication. All datasets used are publicly available.

\item {\bf Experimental setting/details}
    \item[] Question: Does the paper specify all the training and test details (e.g., data splits, hyperparameters, how they were chosen, type of optimizer, etc.) necessary to understand the results?
    \item[] Answer: \answerYes{}
    \item[] Justification: All hyperparameters are specified: temperature (0.1), seed (42), ToT branching factor ($b=3$), ToT depth ($d=3$), NoT max steps (20), NoT min steps (3), convergence threshold (0.8), controller weights ($w_u=0.4$, $w_d=0.4$, $w_c=0.2$).

\item {\bf Experiment statistical significance}
    \item[] Question: Does the paper report error bars suitably and correctly defined or other appropriate information about the statistical significance of the experiments?
    \item[] Answer: \answerNo{}
    \item[] Justification: We use a single seed with near-deterministic temperature (0.1). We acknowledge this limitation in Section~\ref{sec:limitations} and note that multi-seed evaluation with variance estimates is needed for stronger conclusions.

\item {\bf Experiments compute resources}
    \item[] Question: For each experiment, does the paper provide sufficient information on the computer resources (type of compute workers, memory, time of execution) needed to reproduce the experiments?
    \item[] Answer: \answerYes{}
    \item[] Justification: All experiments use API calls to GPT-4o-mini. Token counts per instance are reported in Table~\ref{tab:rq3}. Total API cost for all experiments was around 200 USD.

\item {\bf Code of ethics}
    \item[] Question: Does the research conducted in the paper conform, in every respect, with the NeurIPS Code of Ethics \url{https://neurips.cc/public/EthicsGuidelines}?
    \item[] Answer: \answerYes{}
    \item[] Justification: This work studies reasoning frameworks for language models using publicly available benchmarks. No human subjects, private data, or potentially harmful applications are involved.

\item {\bf Broader impacts}
    \item[] Question: Does the paper discuss both potential positive societal impacts and negative societal impacts of the work performed?
    \item[] Answer: \answerYes{}
    \item[] Justification: Improved reasoning frameworks can benefit applications requiring complex decision-making. We do not foresee direct negative societal impacts from this methodological research.

\item {\bf Safeguards}
    \item[] Question: Does the paper describe safeguards that have been put in place for responsible release of data or models that have a high risk for misuse (e.g., pretrained language models, image generators, or scraped datasets)?
    \item[] Answer: \answerNA{}
    \item[] Justification: This work proposes a reasoning framework, not a new model or dataset. No high-risk assets are released.

\item {\bf Licenses for existing assets}
    \item[] Question: Are the creators or original owners of assets (e.g., code, data, models), used in the paper, properly credited and are the license and terms of use explicitly mentioned and properly respected?
    \item[] Answer: \answerYes{}
    \item[] Justification: All datasets (GSM8K, HotpotQA, ProofWriter) are properly cited. The model (GPT-4o-mini) is accessed through OpenAI's commercial API.

\item {\bf New assets}
    \item[] Question: Are new assets introduced in the paper well documented and is the documentation provided alongside the assets?
    \item[] Answer: \answerYes{}
    \item[] Justification: The NoT framework code and all experimental results will be released with documentation upon publication.

\item {\bf Crowdsourcing and research with human subjects}
    \item[] Question: For crowdsourcing experiments and research with human subjects, does the paper include the full text of instructions given to participants and screenshots, if applicable, as well as details about compensation (if any)?
    \item[] Answer: \answerNA{}
    \item[] Justification: No crowdsourcing or human subjects are involved.

\item {\bf Institutional review board (IRB) approvals or equivalent for research with human subjects}
    \item[] Question: Does the paper describe potential risks incurred by study participants, whether such risks were disclosed to the subjects, and whether Institutional Review Board (IRB) approvals (or an equivalent approval/review based on the requirements of your country or institution) were obtained?
    \item[] Answer: \answerNA{}
    \item[] Justification: No human subjects are involved.

\item {\bf Declaration of LLM usage}
    \item[] Question: Does the paper describe the usage of LLMs if it is an important, original, or non-standard component of the core methods in this research? Note that if the LLM is used only for writing, editing, or formatting purposes and does not impact the core methodology, scientific rigorousness, or originality of the research, declaration is not required.
    \item[] Answer: \answerYes{}
    \item[] Justification: LLMs (GPT-4o-mini) are central to the methodology: they serve as the reasoning engine for all three methods (CoT, ToT, NoT), as the evaluator for LLM-as-Judge, and for self-generated heuristic generation. This usage is fully described in Sections~\ref{sec:framework} and~\ref{sec:eval_metrics}.

\end{enumerate}

\end{document}